# Multi-task deep learning for large-scale building detail extraction from high-resolution satellite imagery

Zhen Qian [a,b,c], Min Chen [a,b,c,*], Zhuo Sun [a,b,c], Fan Zhang [d], Qingsong Xu [e], Jinzhao Guo [a,b,c], Zhiwei Xie [a,b,c,f], Zhixin Zhang [g]

[a] *Key Laboratory of Virtual Geographic Environment (Ministry of Education of PRC), Nanjing Normal University, Nanjing, 210023, China*

[b] *State Key Laboratory Cultivation Base of Geographical Environment Evolution, Nanjing, 210023, China*

[c] *Jiangsu Center for Collaborative Innovation in Geographical Information Resource Development and Application, Nanjing, 210023, China*

[d] *Institute of Remote Sensing and Geographical Information System, School of Earth and Space Sciences, Peking University*

[e] *Department of Aerospace and Geodesy, Data Science in Earth Observation, Technical University of Munich, Arcisstraße 21, Munich, 80333, Bavaria, Germany*

[f] *School of Transportation and Geomatics Engineering, Shenyang Jianzhu University, Shenyang, China*

[g] *College of Geography & Marine, Nanjing University, Nanjing, PO Box 2100913, P.R. China*

* *Corresponding author: chenmin0902@163.com*

## Highlights

- A multi-task network extracting building details from satellite images is proposed.
- Sampling with dual objectives for representative sample selection is designed.
- Representative samples boost prediction accuracy without restructuring network.
- Comprehensive experiments validate the method's accuracy and efficiency.
- Large-scale applications prove generalization, and generated datasets are released.

## Abstract

Understanding urban dynamics and promoting sustainable development requires comprehensive insights about buildings, including their spatial locations, rooftop configurations, physical forms, and urban functions. While geospatial artificial intelligence has advanced the extraction of such details from Earth observational data, existing methods often suffer from computational inefficiencies and inconsistencies when compiling unified building-related datasets for practical applications. To bridge this gap, we introduce the Multi-task Building Refiner (MT-BR), an adaptable neural network tailored for simultaneous extraction of spatial and attributional building details from high-resolution satellite imagery, exemplified by building rooftops, urban functional types, and roof architectural types. Notably, MT-BR can be fine-tuned to incorporate additional building details, extending its applicability. For large-scale applications, we devise a novel spatial sampling scheme that strategically selects limited but representative image samples. This process optimizes both the spatial distribution of samples and the urban environmental characteristics they contain, thus enhancing extraction effectiveness while curtailing data preparation expenditures. We further enhance MT-BR's predictive performance and generalization capabilities through the integration of advanced augmentation techniques. Our quantitative results highlight the efficacy of the proposed methods. Specifically, networks trained with datasets curated via our sampling method demonstrate improved predictive accuracy relative to those using alternative sampling approaches, with no alterations to network architecture. Moreover, MT-BR consistently outperforms other state-of-the-art methods in extracting building details across various metrics, achieving gains between 2% and 20%, with a further 5% improvement when augmentation strategies were adopted. The real-world practicality of our approaches is also demonstrated in an application across Shanghai, generating a unified dataset that encompasses both the spatial and attributional details of buildings. This endeavor also underscores our methodology's potential contribution to urban studies and sustainable development initiatives.

## 1. Introduction

Buildings are not only vital components of urban infrastructure but also represent markers of a city's historical, cultural, and developmental trajectory (Mısırlısoy and Günçe, 2016). Furthermore, the layout and design of buildings are closely tied to the well-being and quality of life of urban inhabitants (Altomonte et al., 2020).

Comprehensive details about buildings, including their spatial locations, rooftop configurations, physical forms, and urban functions, are essential for accurate urban dynamics modeling (Mohajeri et al., 2018; Zhu et al., 2023b). These details are also crucial for guiding decision-making in urban sustainable development, such as leveraging solar energy in the transition towards urban sustainability (Ali et al., 2020; Z. Zhang et al., 2022).

The advent of technological advancements in geoinformatics and remote sensing has ushered in a new era of methodologies for gathering information about building infrastructure (Shi et al., 2020; Zhu et al., 2023b). Traditional data collection methods, which primarily depended on manual interpretation and on-site surveys, have been increasingly replaced by automated, data-driven techniques (Liu et al., 2019). For instance, computer vision methods based on data-driven paradigms have been instrumental in extracting building-related information from Earth observation data sources, such as high-resolution satellite imagery (Wang et al., 2014), digital surface models (Weidner and Förstner, 1995), and light detection and ranging data (Du et al., 2017). In the last decade, the surge in the volume of observational data, along with remarkable strides in computational capabilities, has catalyzed the rise of deep learning techniques (Chen et al., 2023; LeCun et al., 2015). Owing to the ability of deep neural networks to decipher intricate semantic features from large and diverse datasets, the efficacy of computer vision techniques has witnessed significant augmentation (Voulodimos et al., 2018; Xu et al., 2023; F. Zhang et al., 2023). Numerous deep learning-based applications have been developed to acquire a variety of building details essential for advanced urban sustainable development (Cao and Huang, 2021; Z. Zhang et al., 2022).

However, existing building extraction applications primarily concentrate on detecting specific building information from observational datasets, processes generally categorized as single-task operations. While some studies have explored multi-task methodologies, they often emphasize incorporating auxiliary tasks, such as edge detection, to enhance the accuracy of primary tasks like building segmentation (Guo et al., 2021a). This singular focus introduces two technical challenges. Firstly, the straightforward application of existing single-task methods becomes less efficient on larger geographical scales, owing to escalating computational demands and labor costs of training and deploying individual methods to extract varied building details. Secondly, the independently produced datasets often adhere to specific standards, which include data formats and geographic coordinate systems. This strict adherence poses challenges when aiming to merge them into a unified dataset (Qian et al., 2022a). Attempts to address these inconsistencies through post-processing could add more complexity and uncertainty, especially for those in data processing roles (Volk et al., 2014).

Given the aforementioned challenges, multi-task deep learning—integrating deep neural networks with multi-task learning—emerges as a viable solution. Multi-task networks excel at simultaneously processing and refining multiple learning tasks, leveraging shared information across these tasks (Zhang and Yang, 2018). In relation to extracting varied building details from Earth observation data, these networks not only enhance processing efficiency but also improve prediction accuracy (Zhou et al., 2023). Additionally, a significant benefit of multi-task networks is their capacity to produce datasets with consistent data formats and geographic references, ensuring alignment with the standards of the source observational data. Such consistency addresses the data inconsistency challenges inherent to single-task approaches, enhancing the relevance and utility of the extracted building details in wider applications.

Moreover, empirical studies suggest that the diverse physical characteristics of buildings manifest as distinct spatial and spectral patterns in remote sensing images (Qian et al., 2022b). Such diversity introduces complexity to building identification via deep learning, especially when confronting issues like class imbalance (Yang et al., 2023; Zhu et al., 2023a). Much of the current research is geared towards devising advanced methods to surmount these challenges, often leaning on established benchmark datasets (Ji et al., 2018; Maggiori et al., 2017). However, this reliance can inadvertently neglect the pivotal role of data preparation and the necessity of selecting representative samples for effective model training in real-world scenarios. Furthermore, without meticulous sample selection, samples may contain redundant information, as nearby buildings often exhibit similar patterns. This redundancy not only increases annotation costs but also has the potential to cause overfitting of deep learning models. Spatial sampling is advocated as an effective strategy to select representative samples, thereby enhancing the training of neural networks. Previous research corroborates its efficacy in extracting building footprints from high-resolution satellite imagery (Sun et al., 2022; Z. Zhang et al., 2022; Zhong et al., 2021). These studies emphasize the "representativeness" of samples, a concept rooted in geostatistics (Zhang and Zhu, 2019), to assess sample quality. Leveraging representative samples offers two main benefits. Firstly, they provide a diverse range of samples, addressing class imbalance and improving model accuracy and generalizability (Wen et al., 2022). Secondly, a few representative samples can be as effective as a large amount of common samples for model training, thus eliminating annotation and computing costs (Sun et al., 2022). Nonetheless, current spatial sampling methodologies, which adopt stratification strategies or constraints on sample spatial distribution, may not sufficiently capture the nuanced urban environmental characteristics inherent in samples. This oversight might forego potentially valuable insights essential for optimal representative sample selection.

In this study, we present the Multi-task Building Refiner (MT-BR), a multi-task oriented neural network designed to concurrently extract spatial and attributional building details from high-resolution satellite imagery. Specifically, we exemplify building rooftops as spatial details, while taking building's urban functional type and roof architectural type as examples for the attributional aspects. These selected elements play a pivotal role in aiding the urban transition towards sustainable renewable energy and in assessing the impacts of climate change (Z. Zhang et al., 2023). Enhanced with a suite of augmentation techniques, MT-BR excels in discerning a variety of building entities within complex urban environmental scenes. To facilitate extraction of building information across vast geographical expanses, we devised a novel sampling scheme. This method incorporates dual optimization objectives into stratified sampling, considering both the spatial distribution of image samples and the nuanced urban environmental characteristics. This methodology emphasizes the selection of limited but representative samples for training MT-BR, with the goal of enhancing predictive performance while minimizing data preparation costs. The primary contributions of this study can be summarized as follows:

(1) We propose a spatial sampling scheme designed to select limited yet representative samples during the data preparation phase. This approach prioritizes dual objectives related to the spatial distribution of image samples and the nuanced urban environmental characteristics they represent. An enhanced simulated annealing algorithm optimizes these objectives. As a result, the proposed sampling method promotes building information extraction across broad geographical areas while reducing expenses.

(2) We present MT-BR, a multi-task network, crafted for the simultaneous extraction of various building details from high-resolution satellite imagery in an end-to-end manner. The adaptable architecture of MT-BR allows for easy extensions to cater to additional building detail extraction needs. Augmentation techniques are integrated to boost MT-BR's predictive capabilities and enhance its generalization for widespread applications.

(3) Our methods prove effective in large-scale applications, yielding a comprehensive dataset that includes various building details and is available for public access.

The structure of this paper is as follows: Section 2 reviews related work. Section 3 elucidates the methodology we advocate. Section 4 describes the essential materials and outlines our experimental framework. Section 5 presents and discusses our experimental findings. Finally, Section 6 offers our conclusions.

## 2. Related work

### *2.1 Multi-task networks for building detail extraction*

Multi-task networks harness the power of shared representations across individual tasks, utilizing both hard and soft parameter sharing mechanisms (Ruder, 2017). This approach yields three pivotal benefits: firstly, multi-task networks typically have a smaller size compared to the combined sizes of several distinct single-task networks, leading to more efficient storage utilization (Rago et al., 2020). Secondly, thanks to shared feature maps, these networks deliver faster inference, sidestepping the redundancy of calculations for each separate task (Luvizon et al., 2020). Thirdly, the specialized branches tailored for different learning tasks have the potential to enrich the information they process. Notably, integrating a parameter regularization mechanism within these networks can substantially boost the accuracy and generalizability of their predictions (Liu et al., 2015).

Recent studies have explored the several advantages of multi-task networks, with a particular focus on the appeal of parameter regularization, in order to enhance the effectiveness of extracting building information from observational datasets. Many of these studies intefgrate auxiliary tasks to bolster the performance of the main task. For instance, by folding in auxiliary tasks such as edge detection (Guo et al., 2021b; Yin et al., 2022), and frequency and spatial feature representation (Hui et al., 2018; Yu et al., 2022), the prediction accuracy for the core task—identifying building footprints—receives a noticeable uplift. Conversely, tasks like building footprint extraction can also be employed as auxiliary mechanisms, enhancing primary tasks like detecting changes in buildings (Sun et al., 2020). While some studies have explored the extraction of various building details using multi-task networks, the spotlight has largely been on spatial details like building footprints and edges; the attributional information is frequently overlooked.

### *2.2 Spatial sampling scheme*

Acknowledging the spatial dependence and heterogeneity inherent in spatial data, numerous geostatistical studies have evolved beyond traditional random sampling. These studies adopt auxiliary data and existing knowledge to enhance the representativeness of their samples. For instance, Wang et al. (2016) proposed an advanced approach for spatial clustering stratification, which encompasses additional data types like land use and land cover (LULC), administrative boundaries, and expert insights on local areas. Drawing on these geostatistical principles, both Zhong et al. (2021) and Zhang et al. (2022) employed a stratification technique. They divided urban areas into strata based on different levels of built-up density, with the anticipation that strata with high built-up density would yield samples rich in building features. In contrast, strata with lower built-up densities are more likely to generate a higher volume of negative samples, or images devoid of buildings. This strategy is instrumental in mitigating the challenges of foreground-background imbalance encountered in building segmentation from high-resolution satellite imagery.

However, existing sampling methods within various strata largely depend on random mechanisms, often neglecting the spatial and environmental characteristics of the samples. Sun et al. (2022) advanced this approach by integrating spatial optimization within each stratum, with the goal of maximizing the average distance between samples. Their sampling scheme proved to generate more representative image samples compared to those from naive random sampling and stratified sampling, with building segmentation models trained using their dataset achieving superior performance. This distance optimization strategy received additional validation from Liu et al. (2023) in a geomorphological context, further emphasizing the efficacy of such refined spatial sampling. Despite considering the spatial qualities of the samples, this method falls short in adequately incorporating their environmental characteristics during the sampling stage. This limitation highlights an opportunity for enhancement, potentially achievable through the inclusion of more detailed semantic information in the sampling process.

## 3. Methodology

### *3.1 Overview*

This study endeavors to simultaneously extract spatial and attributional building details from high-resolution satellite imagery across extensive geographic expanses efficiently and cost-effectively. To accomplish this, we commence with the procurement of image samples characterized by high representativeness, crucial for the efficacy of subsequent model training. The workflows for this data preparation are depicted in Fig. 1. Our initial step involves overlaying a grid on the study area, with each grid cell considered a potential sampling unit. We then utilize land use and land cover (LULC) data to ascertain the proportion of built-up areas within each grid cell, while points of interest (POIs) help determine the intricate environmental characteristics — specifically, the mixed-use levels at the grid scale. Following the methodology proposed by Zhang et al. (2022) and Sun et al. (2022), we execute a stratification of the grid cells into building-dense and building-sparse categories, based on the built-up proportion. Within this stratified framework, we introduce two optimization objectives that consider both the spatial distribution of samples and their urban environmental characteristics within each stratum. These objectives specifically aim to 1) maximize the average distance between each sample; and 2) maximize the mixed-use levels represented by each sample, which are addressed through an enhanced simulated annealing algorithm. The optimal samples identified are then utilized to extract specific patches from the high-resolution satellite imagery, corresponding to their grid-coordinated locations. These image samples are subsequently annotated manually to delineate rooftops and architectural roof types, while urban functional types are derived using areas of interest (AOIs). The curated datasets are finally apportioned into distinct sets for training, validation, and testing.

Subsequent to these preliminary stages, we present the MT-BR, architected with multiple branches diverging from shared feature maps. Each branch is purposefully designed to extract specific aspects of building information and retains the flexibility for future expansion to accommodate additional building details. The MT-BR incorporates deformable convolutional networks to detect building entities of varying scales and geometries. Moreover, the MT-BR operates under a strategically formulated loss function, which not only guarantees concentrated convergence during training with the selected optimal samples but also establishes a balancing mechanism for the simultaneous learning of heterogeneous building details. To further enhance the MT-BR's prediction performance, we integrate a range of augmentation strategies, including the adoption of advanced backbone structures, ensemble learning techniques, and post-processing procedures. The model's performance is rigorously evaluated against a comprehensive suite of criteria, designed to provide an unbiased assessment of its predictive accuracy and efficiency across diverse building details. Finally, we apply our refined approaches throughout Shanghai, employing geographic information techniques to enhance the quality of the datasets produced.

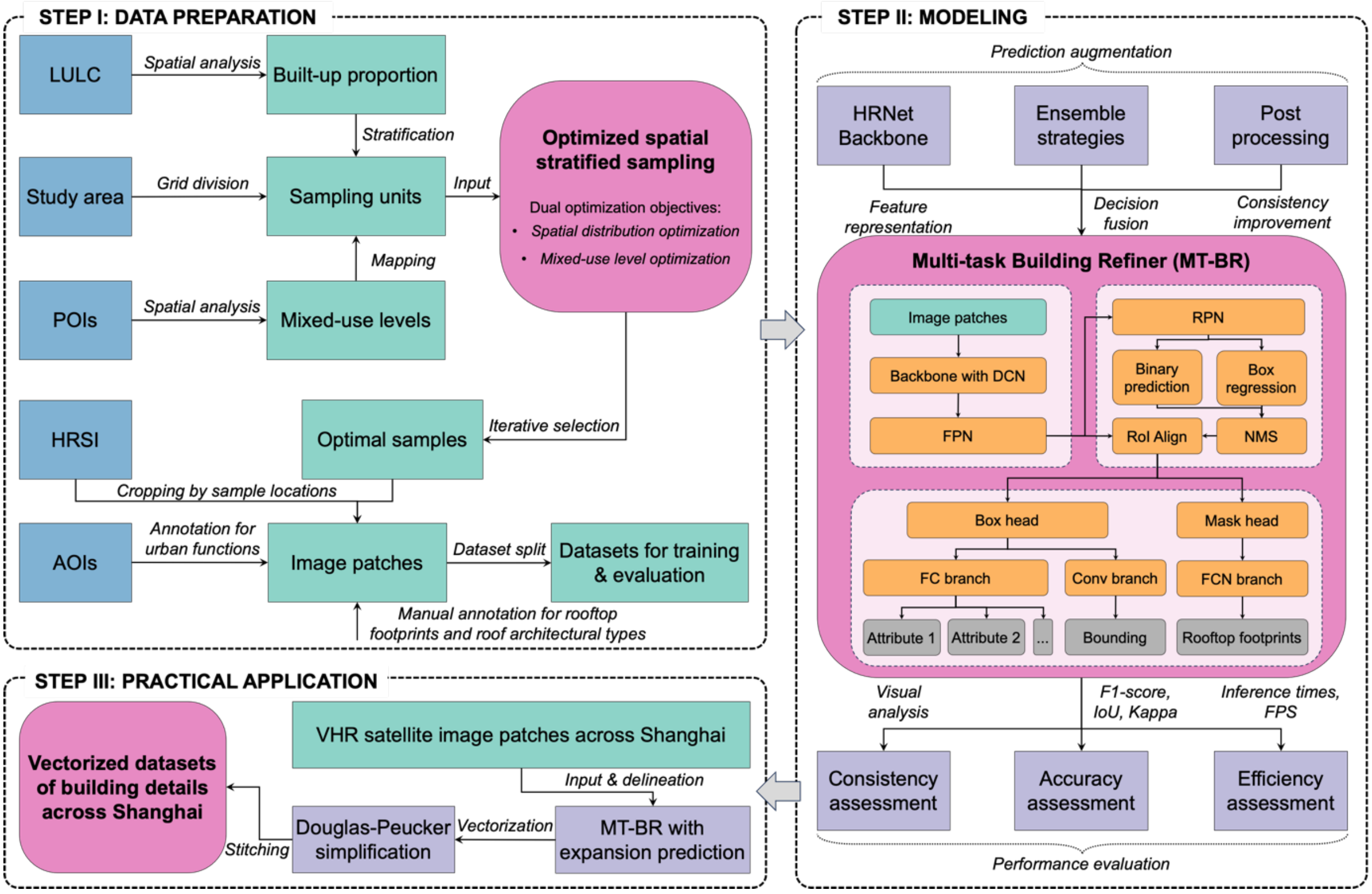

**Fig. 1.** Flowchart of the proposed methods. The process is divided into three phases: 1) preparation of representative datasets, 2) development of a multi-task deep learning network, and 3) large-scale application of the proposed methods for building identification. The segments highlighted in pink indicate the three primary contributions of our study. LULC is land use and land cover data. POIs are points of interest data. HRSI is high-resolution satellite imagery. AOIs are areas of interest data. DCN is deformable convolutional network. FPN is feature pyramid network. RPN is region proposal network. RoI is region of interest. NMS is non-maximum suppression. FC is fully connected operation. Conv is convolutional operation. FCN is fully convolutional network.

### *3.2 Optimized spatial stratified sampling*

#### *3.2.1 Mixed-use levels of urban environments*

Mixed-use levels provide a detailed representation of urban environmental features, particularly highlighting the abundance and diversity of building entities. Various methodologies, encompassing the application of Shannon entropy, have been devised to measure the mixed-use level. Nonetheless, such indices exhibit inherent limitations. Specifically, entropy principally measures uncertainties associated with building distribution, rather than offering a holistic capture of their diversity. Addressing this gap, Yue et al. (2017) turned to Hill numbers, a conceptual framework grounded in ecological research, and proficiently translated it for urban studies application. Hill numbers provide a consolidated representation of diversity metrics, amalgamating richness, entropy, and the Simpson's index (Jost, 2006). The formulation of Hill numbers, as defined by Yue et al. (2017), is given by:

$$ {}^{q}D \equiv \left( \sum_{i=1}^{s} p_i^q \right)^{1/(1-q)} \tag{1} $$

where $D$ signifies the diversity quantification associated with POIs, $s$ denotes the number of distinct POI categories, while $p_i$ represents the proportion of POIs ascribed to the $i$th category. Additionally, $q$ functions as a diversity order parameter. When $q$ is 0, the metric reverts to a richness index, signifying the number of unique POI categories within a delineated region. In contrast, when $q$ takes the value of 1, it evolves into an exponential of the Shannon entropy, providing insights into the level of orderliness in both POI categories and quantities. Upon reaching $q = 2$, the metric converges to the inverse of the Simpson index, accounting for both the variance in POI types and the relative proportions of disparate POI categories (Yue et al., 2017).

Hill numbers offer a holistic and quantifiable mechanism to delineate both the richness and diversity inherent to buildings. Conforming to the methodology proposed by Yue et al. (2017), our investigation likewise deploys POIs to compute Hill numbers, thereby assessing the mixed-use levels of urban environments. To synthesize a descriptive metric amalgamating the three measures derived from Hill numbers, we execute a standardization and subsequent averaging. This synthesized metric furnishes a quantitative representation of mixed-use levels to inform subsequent sampling procedures:

$$MUL_n = \frac{\sum_{q=0}^{Q} \mathrm{std}({}^{q}D)}{Q+1} \tag{2}$$

where $MUL_n$ signifies a synthesized metric representing the mixed-use level for the $n$th sample, as calculated via Hill numbers, $\mathrm{std}(\cdot)$ indicates the min-max normalization operation, and $Q$ stands for the order of the diversity and abundance, set to 2 in our study.

### *3.2.2 Simulated annealing-based dual-objective optimization*

Moving beyond the approach that prioritizes only the spatial distribution of samples as the optimization objective, we expanded the sampling scheme to integrate the mixed-use levels of urban environment. As a result, the dual optimization objectives now focus on minimizing cost functions related to the inverse of the average nearest neighbor ($Cost_{ANN}$) (Clark and Evans, 1954) and the inverse of the average mixed-use level ($Cost_{AMUL}$). These objectives are articulated as follows:

$$Cost_{ANN} = \left( \frac{\sum_{n=1}^{N} D_n / N}{1 / 2\sqrt{N/A}} \right)^{-1} \tag{3}$$

$$Cost_{AMUL} = \left( \frac{\sum_{n=1}^{N} MUL_n}{N} \right)^{-1} \tag{4}$$

$$\begin{aligned} \min \quad & \{\mathrm{Cost_{ANN}}, Cost_{AMUL}\} \\ s.\mathrm{t.} \quad & N > 0, \\ & A > 0, \\ & D_n \geq 0, \\ & ANN \geq 0, \\ & 0 \leq MUL_n \leq 1 \end{aligned} \tag{5}$$

where, $D_n$ denotes the distance between the $n$th sample and its nearest neighboring sample, $N$ is the total number of samples, and $A$ indicates the area covering all samples.

The simulated annealing algorithm is improved to handle both objectives together. This approach is inspired by the annealing process in metallurgy, a process where a material is heated and then cooled to reduce defects (Rutenbar, 1989). In optimization, simulated annealing uses a similar approach to find an optimal solution in a complex space. This technique embodies a stochastic approach that empowers the algorithm to navigate solution spaces and circumvent local optima. The algorithm operates by iteratively investigating neighboring solutions, and intriguingly, it accepts new solutions even when they might be worse than the current one. This acceptance criterion, see Eqs. (6) and (7), characterized by a decreasing probability over time, introduces an element of randomness (Bertsimas and Tsitsiklis, 1993). It permits the algorithm to occasionally accept suboptimal solutions, as the following formulations, thereby facilitating its ability to transcend local optima and persistently explore the expansive solution space.

$$P(S_{optimal}) = \begin{cases} 1, & \Delta\text{Cost} < 0 \\ \exp\left(\frac{-\Delta\text{Cost}}{T_e}\right), & \Delta\text{Cost} \geq 0 \end{cases} \tag{6}$$

$$T_{e+1} = \alpha \cdot T_e \tag{7}$$

where $P(S_{optimal})$ denotes the probability of accepting the optimal samples chosen during the $e$th iteration, $\Delta\text{Cost}$ represents the change in the cost function between the $(e-1)$th iteration and the $e$th iteration, $T_e$ is the temperature parameter for the $e$th iteration, commencing with the initial temperature when $e = 0$, and $\alpha$ is the decay parameter, often set as a constant value smaller than 1.

Within the enhanced simulated annealing framework, the algorithm adeptly manages two distinct solutions pertaining to different objectives. The stepwise procedures for handling these solutions are outlined as follows:

**Algorithm 1** Enhanced Simulated Annealing for Dual-Objective Optimization

**Input:** $Cost_{ANN}$ and $Cost_{AMUL}$, cost functions.

$S_{total}$, full set of samples; each sample defined by ID ($id_n$), longitude ($X_n$), latitude ($Y_n$), and mixed-use levels ($MUL_n$) attributes.

$N$, desired sampling count.

$A$, sampling area.

$T_0$, initial temperature.

$\alpha$, temperature decay rate.

$T_{tol}$, minimum tolerable temperature.

$E$, maximum iteration count.

**Initialize** $\Delta\text{Cost}_{\text{ANN}} = 0$, $\Delta\text{Cost}_{\text{AMUL}} = 0$, and $\text{e} = 0$;

Randomly select $N$ samples from $S_{total}$ to form the initial sampling solution $S_{optimal}$;

Calculate $Cost_{ANN_e}$ and $Cost_{AMUL_e}$ using Eqs. (3) and (4);

**while** $T_e > T_{tol}$ and $e < E$ **do**

Increment $e$ by 1;

Perturb $S_{optimal}$ as follows:

1. Randomly select a sample from the difference set of $S_{total}$ and $S_{optimal}$ to replace the sample in $S_{optimal}$ with the shortest nearest neighboring distance;
2. Randomly select another sample from the difference set of $S_{total}$ and $S_{optimal}$ to replace the sample in $S_{optimal}$ with the lowest mixed-use level;

Compute $Cost_{ANN_e}$, $Cost_{AMUL_e}$, $\Delta\text{Cost}_{\text{ANN}}$, and $\Delta\text{Cost}_{\text{AMUL}}$ using Eqs. (3) and (4);

Determine $T_e$ and $P(S_{optimal})$ using Eqs. (6) and (7); Note: $P(S_{optimal})$ consists of $P_{ANN}(S_{optimal})$, and $P_{AMUL}(S_{optimal})$;

**if** $P(S_{optimal}) >$ random number between 0 and 1 **then**

Accept the perturbed samples;

**else**

Retain the original sampling solution;

**end if**

**end while**

Update $S_{most_optimal}$ to $S_{optimal}$;

**Output:** $S_{most_optimal}$, the best set of samples obtained.

### *3.3 Multi-task building refiner*

To holistically extract various building details, we propose a multi-task network, named MT-BR. This architecture is essentially an intuitive extension of the Mask R-CNN framework (He et al., 2017). The overall structure of MT-BR is depicted in Fig. 2. MT-BR is characterized by its multiple specialized branches, each meticulously crafted for versatility and scalability. Key to its performance is the integration of deformable convolutional networks, which are adept at identifying buildings that span a range of scales and possess distinct geometrical features. Accompanying this is a tailored loss function, specifically crafted for the demands of multi-task learning. Furthermore, to enhance the model's performance and generalizability, we have infused MT-BR with three strategic augmentation techniques.

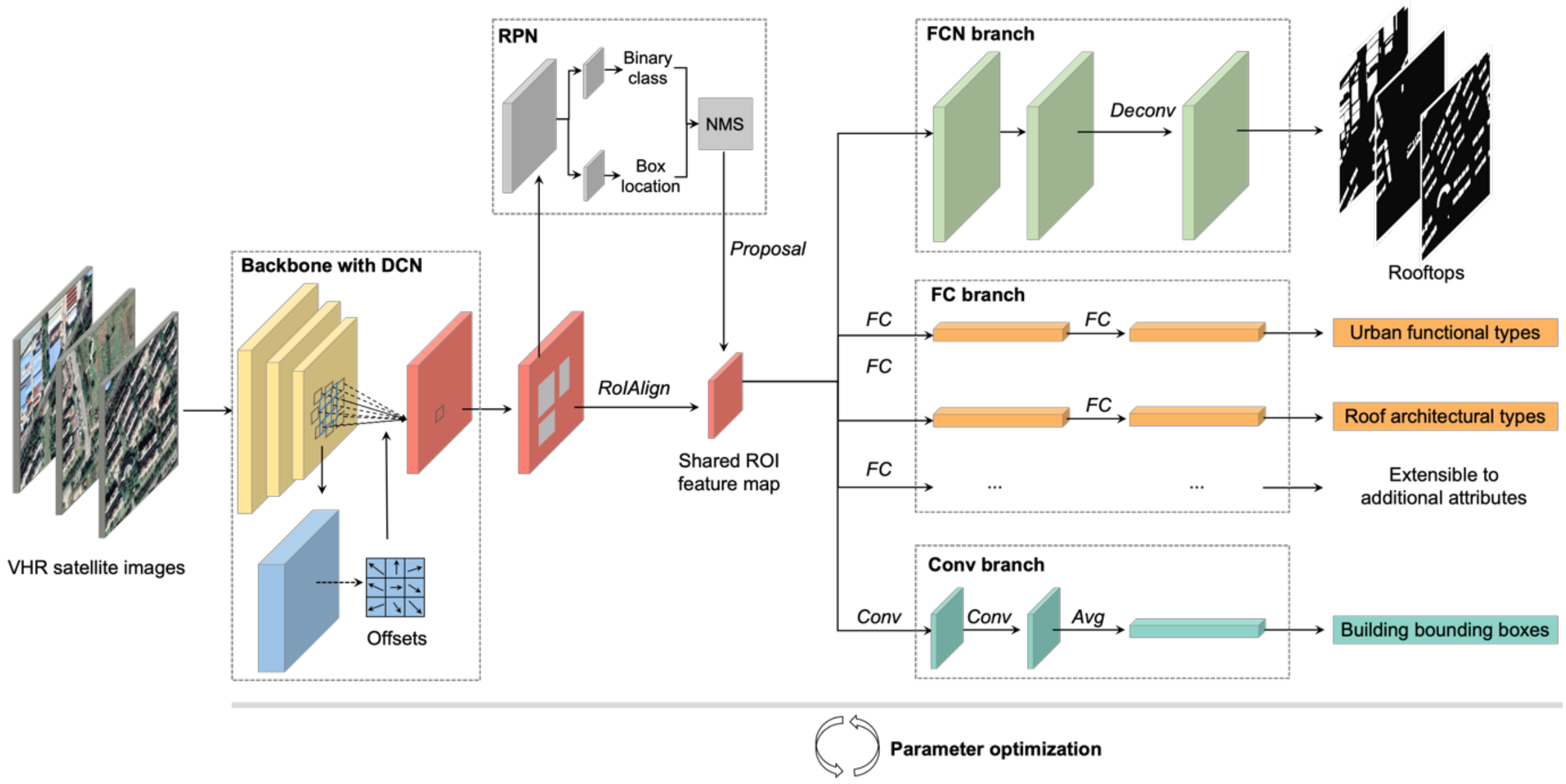

$$\mathcal{L}_{\text{total}} = \frac{1}{n}\left(\mathcal{L}_{cls_{func}} + \mathcal{L}_{cls_{arch}} + \mathcal{L}_{cls_{...}}\right) + \mathcal{L}_{\text{loc}} + \mathcal{L}_{\text{mask}}$$

**Fig. 2.** Overview of the MT-BR architecture. The network utilizes a Deformable Convolutional Network (DCN) to accommodate buildings of various scales and shapes. Its scalable branches, characterized by an integrated use of convolutional (Conv) and fully connected (FC) layers, are crafted within a multi-task learning paradigm, allowing for the concurrent extraction of various building details. The custom-designed loss function facilitates multi-task learning and can be expanded to accommodate additional building information extraction tasks. RPN is region proposal network. RoI is region-of-interest. NMS is non-maximum suppression. FC is full connected operation. Avg is average pooling. Deconv is deconvolutional operation.

### *3.3.1 Multi-branch network architecture*

The MT-BR architecture incorporates several distinct branches, intuitively extended from the Mask R-CNN, to simultaneously extract heterogenous building details. These branches have designated roles: segmenting rooftops, localizing building bounding boxes, classifying roof architectural types, and identifying urban functional types of buildings. The bounding box localization primarily plays a supportive auxiliary role. All these branches operate collaboratively, leveraging shared region-of-interest (RoI) feature maps. Considering the diversity in learning tasks between these branches, especially in object classification and spatial localization, our methodology draws inspiration from Wu et al. (2020). Specifically, to enhance the precision in extracting building details, attributional classification branches utilize fully connected layers, while localization branches are built on convolutional layers. A standout feature of the MT-BR is its flexible multi-branch design. This flexibility allows for the seamless integration of additional convolutional or fully connected branches based on the characteristics of specific learning tasks.

### *3.3.2 Deformable convolutional network*

Traditional convolutional networks, characterized by their fixed spatial sampling locations stemming from regular geometric structures, often struggle to accurately represent targets with diverse shapes, sizes, and orientations. Such diversity is commonly observed in geographic objects present in remote sensing images. To overcome these representation challenges, deformable convolution and deformable RoI pooling were introduced (Dai et al., 2017; Zhu et al., 2019) , granting neural networks the ability to adapt their spatial sampling locations. This adaptability is achieved through the incorporation of learnable offsets, added to the conventional convolution operations as depicted in Eq. (8). Notably, no additional supervision is needed for these offsets. An interesting perspective on deformable convolution is its resemblance to a local attention mechanism, which enables the model to prioritize salient patterns in a localized manner. This attention mechanism enhances the model's capacity to discern contextual information across various scales. Within the context of satellite imagery, the deformable strategies exhibit the capability to discern distinct building objects, minimizing misinterpretations arising from intricate environmental backgrounds (Zhu et al., 2018).

$$y(p_0) = \sum_{p_n \in R} w(p_n) \cdot x(p_0 + p_n + \Delta p_n) \tag{8}$$

where $x$ represents the input feature map, $y$ denotes the output feature map, $p_n$ is the position set for a convolution operation—with $p_n$ belonging to a set that includes positions like $(-1, -1)$, $(-1,0)$, up to $(0,1)$ and $(1,1)$ for a 3 × 3 convolution—and $\Delta p_n$ corresponds to the learnable offset.

### *3.3.3 Loss function*

The primary objective of the loss function in our approach is to facilitate gradient backpropagation for multi-task learning, specifically tailored for extracting various building details. Our methodology extends the original loss function from Mask R-CNN (He et al., 2017), with a modification being the use of the mean to compute the loss for attributional information learning, emphasizing its efficiency and simplicity. This design choice also ensures that the model remains scalable and can efficiently integrate any potential building attributional classification tasks in the future. The multi-task loss function, as detailed in our study, is articulated in the subsequent equations:

$$\mathcal{L}_{\mathrm{cls}} = -{y_{\mathrm{cls}}}^{\mathrm{T}} \log \hat{y}_{\mathrm{cls}} \tag{9}$$

$$\mathcal{L}_{\mathrm{loc}} = \begin{cases} 0.5|y_{\mathrm{loc}} - \hat{y}_{\mathrm{loc}}|^2, & |y_{loc} - \hat{y}_{\mathrm{loc}}| < 1 \\ |y_{\mathrm{loc}} - \hat{y}_{\mathrm{loc}}| - 0.5, & |y_{loc} - \hat{y}_{\mathrm{loc}}| \geq 1 \end{cases} \tag{10}$$

$$\mathcal{L}_{\mathrm{mask}} = -\frac{1}{n^2} \sum_{1 \leq i,j \leq n} y_{\mathrm{mask}_{ij}} \log \hat{y}_{\mathrm{mask}_{ij}} + (1 - y_{\mathrm{mask}_{ij}}) \log(1 - \hat{y}_{\mathrm{mask}_{ij}}) \tag{11}$$

$$\mathcal{L}_{\mathrm{total}} = \frac{1}{2}\left(\mathcal{L}_{\mathrm{cls_{func}}} + \mathcal{L}_{\mathrm{cls}_{arch}}\right) + \mathcal{L}_{\mathrm{loc}} + \mathcal{L}_{\mathrm{mask}} \tag{12}$$

where $\mathcal{L}_{\mathrm{cls}}$ is the cross-entropy loss for classification task (Ren et al., 2015), $\mathcal{L}_{\mathrm{loc}}$ is the smooth-L1 loss for localization task (Ren et al., 2015), $\mathcal{L}_{\mathrm{mask}}$ is the binary cross-entropy loss for segmentation task (He et al., 2017), $y_{\mathrm{cls}}$ and $\hat{y}_{\mathrm{cls}}$ are the ground-truth and predicted labels of RoIs, $y_{loc}$ and $\hat{y}_{\mathrm{loc}}$ are the ground-truth and predicted coordinates of RoIs, $n$ is the size of output images, $y_{\mathrm{mask}_{ij}}$ and $\hat{y}_{\mathrm{mask}_{ij}}$ are the ground-truth and predicted labels for the image cell at position $(i, j)$.

### *3.3.4 Prediction augmentation*

To bolster the prediction accuracy of MT-BR and address potential inconsistencies in its outputs, especially for real-world applications, we have integrated three prediction augmentation strategies.

1) *HRNet Backbone*

The High-Resolution Network (HRNet) offers a solution to the prevalent challenge of diminishing resolution in feature maps as convolution layers proliferate (Sun et al., 2019). This proficiency stems from its innovative architecture, characterized by key features such as multi-scale fusion, resilient skip connections, and consistent high-resolution representation. These features collectively ensure the preservation of vital contextual information from input data, guaranteeing the retention of intricate details, irrespective of the network's depth. Given that extraction tasks necessitate nuanced and sharp feature representations to distinguish diverse building attributes, HRNet's architecture is optimally suited to meet these demands.

2) *Ensemble Strategies*

Integrating deep neural networks with ensemble learning techniques has consistently demonstrated its efficacy in amplifying prediction accuracy and bolstering model generalization (K. Zhang et al., 2022). Two prominent techniques, multi-model decision fusion and test-time augmentation, have been particularly effective in refining post-training results, establishing their utility across a spectrum of applications. In our methodology, we capitalize on ensemble learning by training an array of MT-BRs, each equipped with distinct backbones, drawing from the Bagging concept. This approach that the model ensembles exhibit a wide variance in their predictions, potentially bolstering holistic performance. During the inference process, we adopt a test-time data augmentation strategy, subjecting the data to manipulations such as multi-scale resizing and rotation to enrich input diversity. The predictions derived from these varied inputs are then aggregated to produce consolidated outcomes. By integrating these ensemble techniques, our aim is to harness the collective strengths of multiple models and varied data inputs. This approach seeks to enhance both the prediction accuracy and generalization capability when extracting diverse building details from high-resolution satellite imagery.

3) *Post-processing*

During the inference stage of the MT-BR, multiple detection boxes—each indicating different predicted labels—can overlay the same building entity. This overlap can result in discrepancies within the predicted pixels of a singular building entity. To address this issue, we assess the prevalence of each predicted pixel class for individual building entities. Following this, the most commonly predicted label is designated as the conclusive attributional label for that entity. This process can be mathematically described as:

$$attr_i = \underset{c}{\mathrm{argmax}}\left(\frac{\mathrm{count}(c, C_i)}{|C_i|}\right) \tag{13}$$

where $attr_i$ denotes the definitive attributional label of the $i$th building entity, $C_i$ represents the set of predicted pixel classes for the $i$th entity, $\mathrm{count}(c, C_i)$ is the frequency of class $c$ within $C_i$, and $|C_i|$ corresponds to the aggregate count of $C_i$.

## 4. Experimental preparation

### *4.1 Materials*

#### *4.1.1 Study area*

Shanghai, situated in China's eastern region and depicted in Fig. 3, serves as the study area of this research. Covering an expansive administrative area of approximately 6340 km², Shanghai boasts a population exceeding 24 million, solidifying its position as one of China's most significant and densely populated metropolises. Nationally recognized for its leading role in construction and economic progression, Shanghai's intense urbanization and intricate architectural tapestry render it a fitting subject for our study. The LULC status of Shanghai, displayed on the right side of Fig. 3, was sourced from ESRI in 2022 and possesses a spatial resolution of 10 meters (Karra et al., 2021). This data reveals that built-up regions account for over 60% of Shanghai's total land area. The city's urban layout is distinguished by its complex patterns and varied functional zones. Extracting building details from high-resolution satellite imagery within such a multifaceted urban setting presents formidable challenges. Nevertheless, accomplishing this task in Shanghai holds significant value, with potential implications for enhancing urban sustainability initiatives, refining management strategies, and enriching the repository of urban geographic information.

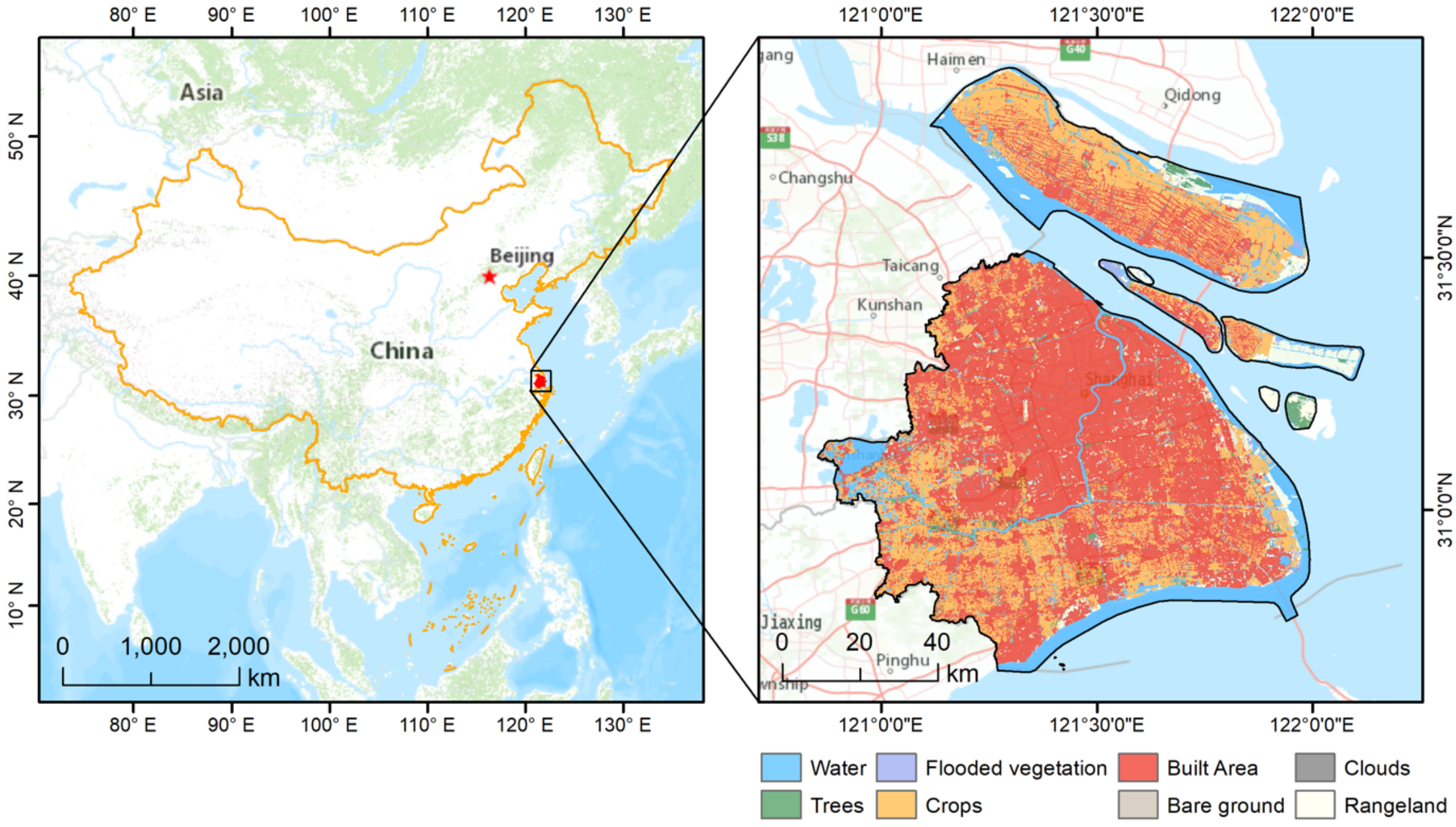

**Fig. 3.** Geographic representation of Shanghai's location and its land use and land cover distribution.

### *4.1.2 High-resolution satellite imagery*

High-resolution satellite imagery was obtained from satellites like the WorldView-2/3 series, GeoEye-1, SkySat, and Pleiades in 2022. These images combine both panchromatic and multispectral data captured at the same time by the satellite and are normalized to a 0.5 m resolution (Guo et al., 2023). An illustrative example is presented in Fig. 4 (a).

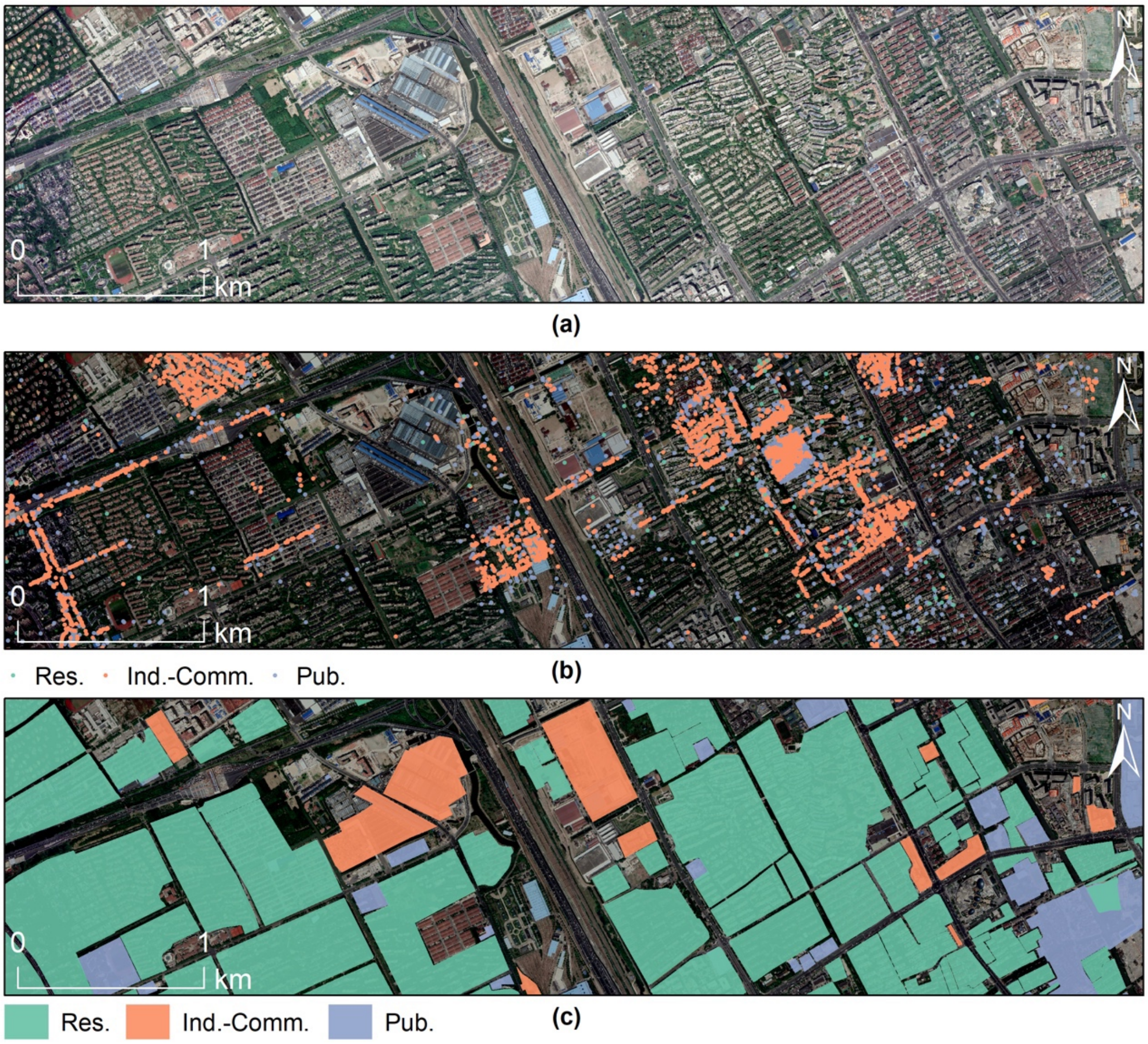

**Fig. 4.** Datasets used in this study. (a) High-resolution satellite image. (b) POIs. (c) AOIs.

#### *4.1.3 Social sensing data*

The social sensing data utilized in this study encompasses POIs and AOIs. These datasets not only identify geographic object locations but also provide insights into their functional roles. Moreover, they offer a glimpse into the socio-economic context and the dominant urban environmental characteristics of the area. Through open application programming interfaces from AMap, we have aggregated an extensive dataset for Shanghai as of 2022, which includes 1.5 million POIs and 2.8 thousand AOIs. The information captured within these POIs and AOIs spans various attributes, including title, type, province, city, address, longitude, and latitude. Due to the distinct spatial scales at which POIs and AOIs are depicted, disparities in the urban functions they depict may arise. Aligned with our research objectives, we undertook a reclassification of the functional types of the collated POIs and AOIs. Furthermore, we streamlined them into three primary categories: residential, industrial-commercial, ad public services, as shown in Fig.4 (b) and Fig.4 (c). This classification adhered to AMap's categorization guidelines and insights from Zhang et al. (Z. Zhang et al., 2023).

### *4.2 Description and assessment of various building details*

#### *4.2.1 Building spatial information*

Our study defines rooftops as a representation of building spatial information, visualized as a two-dimensional projection of building roofs (see Fig. 5). A technical challenge emerges due to the disparity in size between original satellite image and the typical image patch size fed into deep neural networks. Such mismatches can lead to scenarios where a single building is split across multiple image patches, causing alignment issues upon stitching. Moreover, the precision of pixel-level building roof recognition can sometimes yield rough edges. These challenges can be mitigated using techniques such as expansion prediction and vector simplification (Z. Zhang et al., 2022).

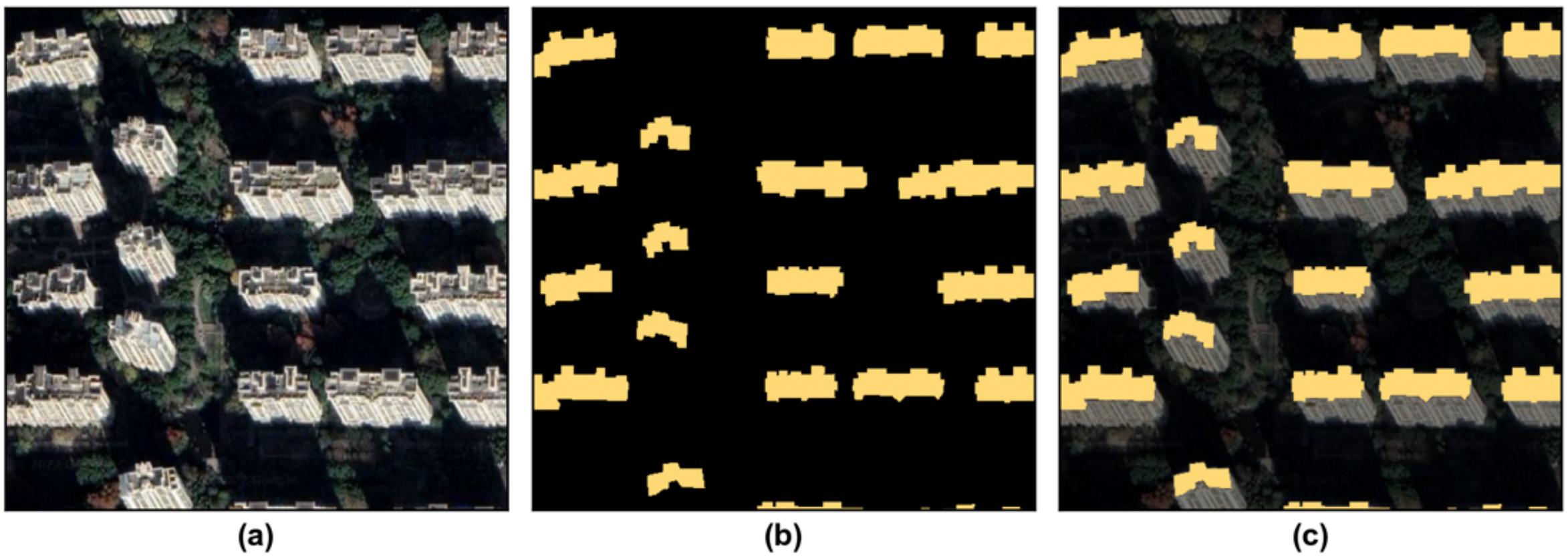

**Fig. 5.** Illustration of building rooftops. (a) High-resolution satellite image. (b) Annotated rooftops. (c) High-resolution satellite image overlaid with rooftop annotations.

To gauge the efficacy of delineating rooftops, we turn to established pixel-level evaluation metrics: the F1-score and the Intersection over Union (IoU) (Li et al., 2019). Prior to computing these metrics, foundational metrics like precision and recall need to be derived from the confusion matrix. Their respective equations are provided below:

$$\text{Precision} = \frac{\text{TP}}{\text{TP} + \text{FP}} \tag{14}$$

$$\text{Recall} = \frac{\text{TP}}{\text{TP} + \text{FN}} \tag{15}$$

$$\mathrm{F1} = \frac{2 * \mathrm{Precision} * \mathrm{Recall}}{\mathrm{Precision} + \mathrm{Recall}} \quad (16)$$

$$\mathrm{IoU} = \frac{\mathrm{TP}}{\mathrm{TP} + \mathrm{FN} + \mathrm{FP}} \quad (17)$$

where TP represents true positive predictions, FP stands for false positive predictions, and FN denotes false negative predictions.

#### *4.2.2 Building attributional information*

The attributional information of buildings covers the urban functional types and roof architectural types in this study. Urban functional types encompass the fundamental responsibilities that buildings undertake in urban environments, such as residential, industrial, commercial, and public utilities. Past research has delineated urban functional types at a more macroscopic level, focusing on territorial units like contiguous grids or adjacent blocks (Lu et al., 2022; Qian et al., 2020). Our endeavor, however, targets the identification of specific functional categories at the granularity of individual building entities. This shift in scale considerably amplifies the intricacy of the modeling process relative to earlier methodologies. Moreover, based on empirical observations, it has been discovered that buildings with similar designs in specific places, such as industrial zones, tend to be classified as either commercial or industrial. Hence, our examination centers on three discrete classifications: residential, industrial-commercial, and public service categories, as shown in Fig.4 (c).

Roof architectural types shed light on a city's architectural progression and cultural heritage, crucial for cultural preservation and urban planning research (Sun et al., 2017; Wang et al., 2022). In this study, Shanghai's roof architectural types are broadly divided into four main categories: flat, gable, hip, and complex, following the classifications by Mohajeri et al. (2018). These categories are visually presented in Fig. 6. Complex roofs encompass a variety of designs, combining features from multiple basic architectural types or presenting distinct designs indicative of a building's style.

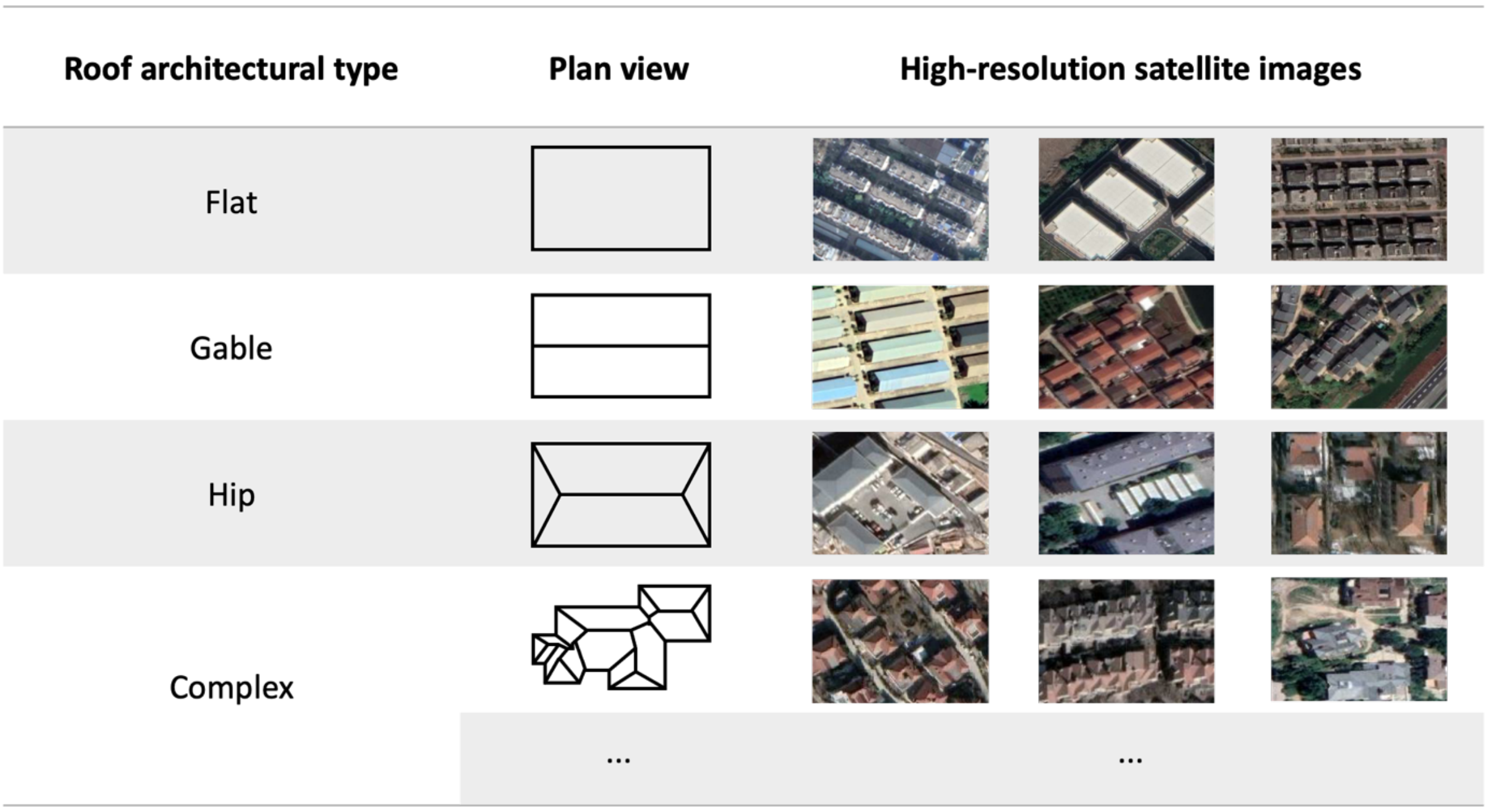

**Fig. 6.** Depiction of various roof architectural types.

To assess the efficacy of our approach in extracting building attributional information, we adopt the Kappa coefficient, a metric that has garnered widespread adoption in the domain of remote sensing image classification (Huang et al., 2018). The Kappa coefficient oscillates between -1 and 1, with values exceeding 0.6 typically denoting a robust alignment between the derived classification outcomes and the ground truth, thereby vouching for the robustness of the results. The Kappa coefficient, , as defined by Chmura Kraemer et al. (2002), is given as:

$$\text{Kappa} = \frac{p_o - p_e}{1 - p_e} \qquad (18)$$

where $p_o$ represents the observed accuracy, signifying the fraction of accurately classified instances, $p_e$ symbolizes the expected accuracy under random classification scenarios, deduced as the mean of the probabilities associated with each category emanating from the classifier's output.

### *4.3 Experimental setup*

Our experiments are structured around two main components: data sampling and building information extraction. Data sampling is primarily conducted on a personal computer, equipped with an Intel i7-10700K CPU, 64GB RAM, and an NVIDIA GeForce RTX 3090 GPU. The software stack for this phase includes the Windows operating system, the ArcGIS platform, and a Python development environment. In contrast, the building information extraction phase is executed on a robust supercomputing platform that runs on the Ubuntu operating system. The computational framework is fitted with eight NVIDIA GeForce RTX 3090 GPUs and leverages software tools such as the Python environment, PyTorch (Paszke et al., 2019), and MMdetection (Chen et al., 2019b).

To guarantee the integrity and fairness of our experiments, we've adhered to uniform standards across the board. Specifically, the input data size for each model remains fixed at 512 × 512 pixels. Additionally, we've standardized model hyperparameters, such as the backbone network, optimizer, and learning rate, for consistent multi-model comparisons. A detailed description of these hyperparameters is available in Table 1. Acknowledging the stochastic nature in training deep neural networks, we repeated every training and evaluation sessions five times. For transparency, we present our quantitative metric outcomes as mean values accompanied by standard deviations.

**Table 1** Experimental configuration details.

| Hyperparameter | Value |
|---|---|
| Image size | 512 × 512 |
| Backbone | ResNet101 |
| Optimizer | Stochastic gradient descent |
| Learning rate | 0.001 |
| Batch size | 4 |
| Epoch | 108 |

## 5. Experimental results and discussions

### *5.1 Data sampling experiments*

#### *5.1.1 Sampling preparation*

For the collection of image samples that are highly representative for training our model, we overlay a grid on the study area, where each cell measures 1 × 1 km. Each grid cell serves as a distinct sampling unit, and image samples are subsequently extracted from the high-resolution satellite imagery corresponding to the spatial coordinates of optimal units. To determine the urbanization degree within each grid, ESRI LULC data is used to calculate the proportion of built-up areas in every grid, presented in Fig. 7 (a). Subsequently, POIs are harnessed to calculate the mixed-use levels at the grid level, as shown in Fig. 7(b), in line with our proposed method. Spatial patterns of both built-up proportions and mixed-use levels are closely correlated, particularly evident in urbanized districts such as Huangpu, Changning, Jing'an, Xuhui, Yangpu, Hongkou, and Putuo. To differentiate between building-dense and building-sparse strata, we use the lower quartile value of the built-up proportion, established at 0.16, as recommended by Zhang et al. (2022) and Sun et al. (2022). This stratification is illustrated in Fig. 7(c) and Fig. 7(d).

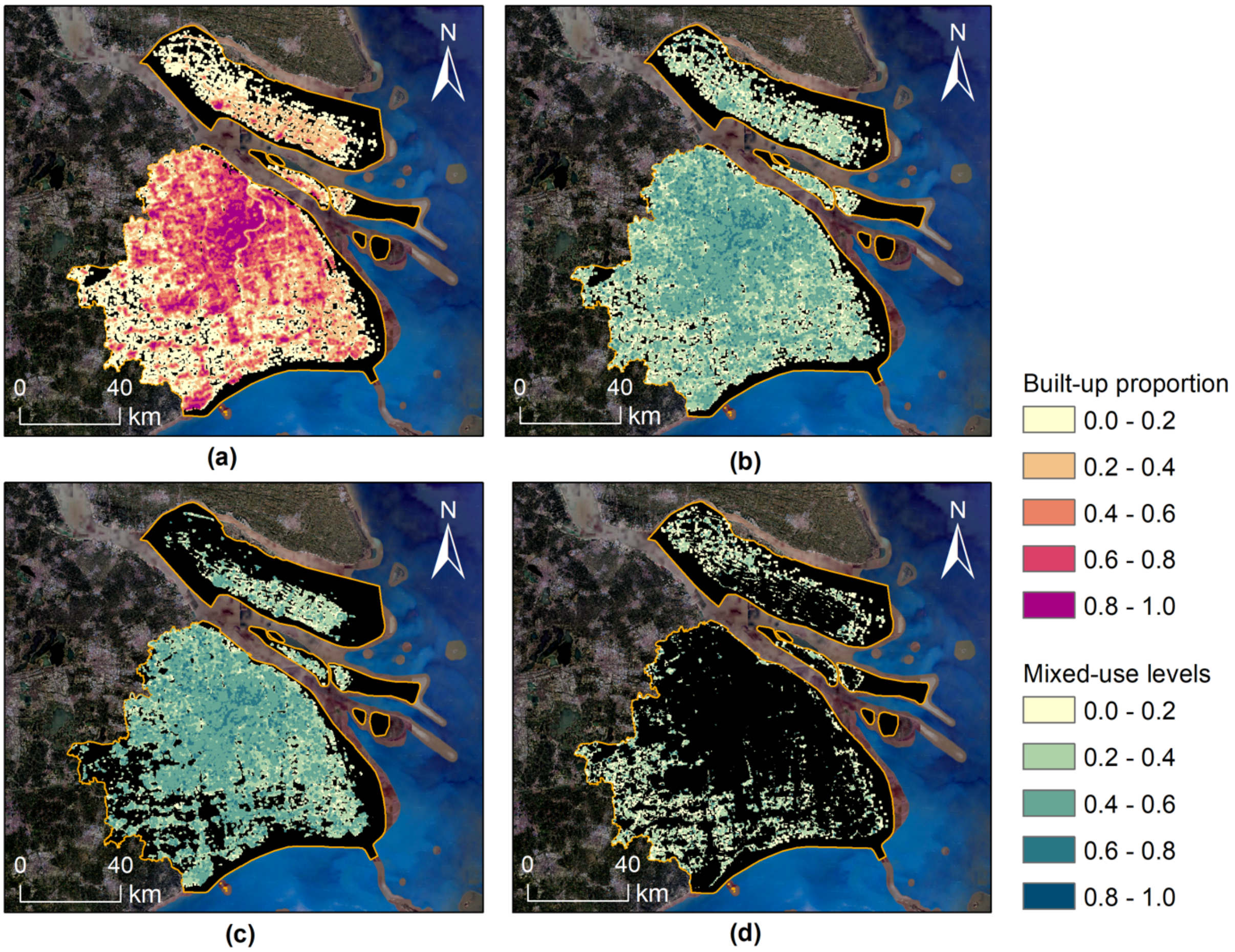

**Fig. 7.** Depictions of built-up proportion and mixed-use levels. (a) Built-up proportion mapping. (b) Mixed-use level mapping. (c) Mixed-use levels within the building-dense stratum. (d) Mixed-use levels within the building-sparse stratum.

### *5.1.2 Sampling results and analysis*

Utilizing the proposed sampling schme with dual-objective optimization, we select a total of 100 sampling units, each spanning an area of 1 × 1 km. The majority, 80 units, are focused on the building-dense stratum, with the remaining 20 targeting the building-sparse stratum. The optimization process, which uses the simulated annealing algorithm, undergo 5000 iterations, with its progression charted in Fig. 8.

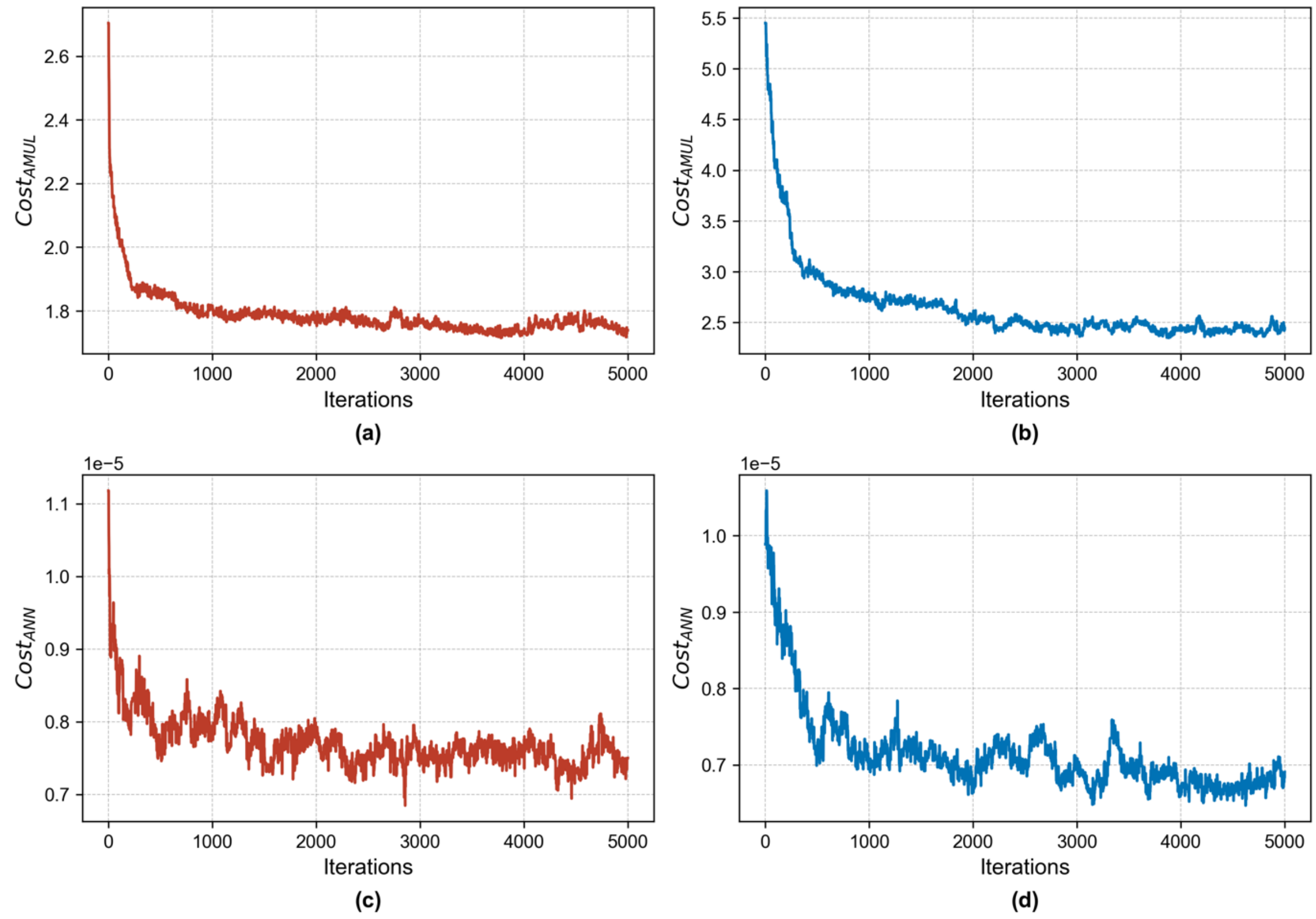

**Fig. 8.** Dual-objective optimization progression. (a) Mixed-use levels optimization within the building-dense stratum. (b) Mixed-use levels optimization within the building-sparse stratum. (c) Spatial distribution optimization within the building-dense stratum. (d) Spatial distribution optimization within the building-sparse stratum.

Convergence in the building-sparse stratum is achieved more rapidly than in the building-dense stratum, due to its smaller area and reduced number of sampling units. Approximately 3000 iterations are needed for the building-sparse stratum, while the building-dense stratum necessitates around 4000 iterations. During the optimization, spatial sample distribution exhibits more fluctuations, while the optimization for mixed-use levels remains relatively stable. This disparity arises from the intertwined interactions during the dual-objective optimization. The spatial locations of these optimal samples are visualized in Fig. 9, which highlights samples spread across different locales, primarily in regions with high mixed-use levels.

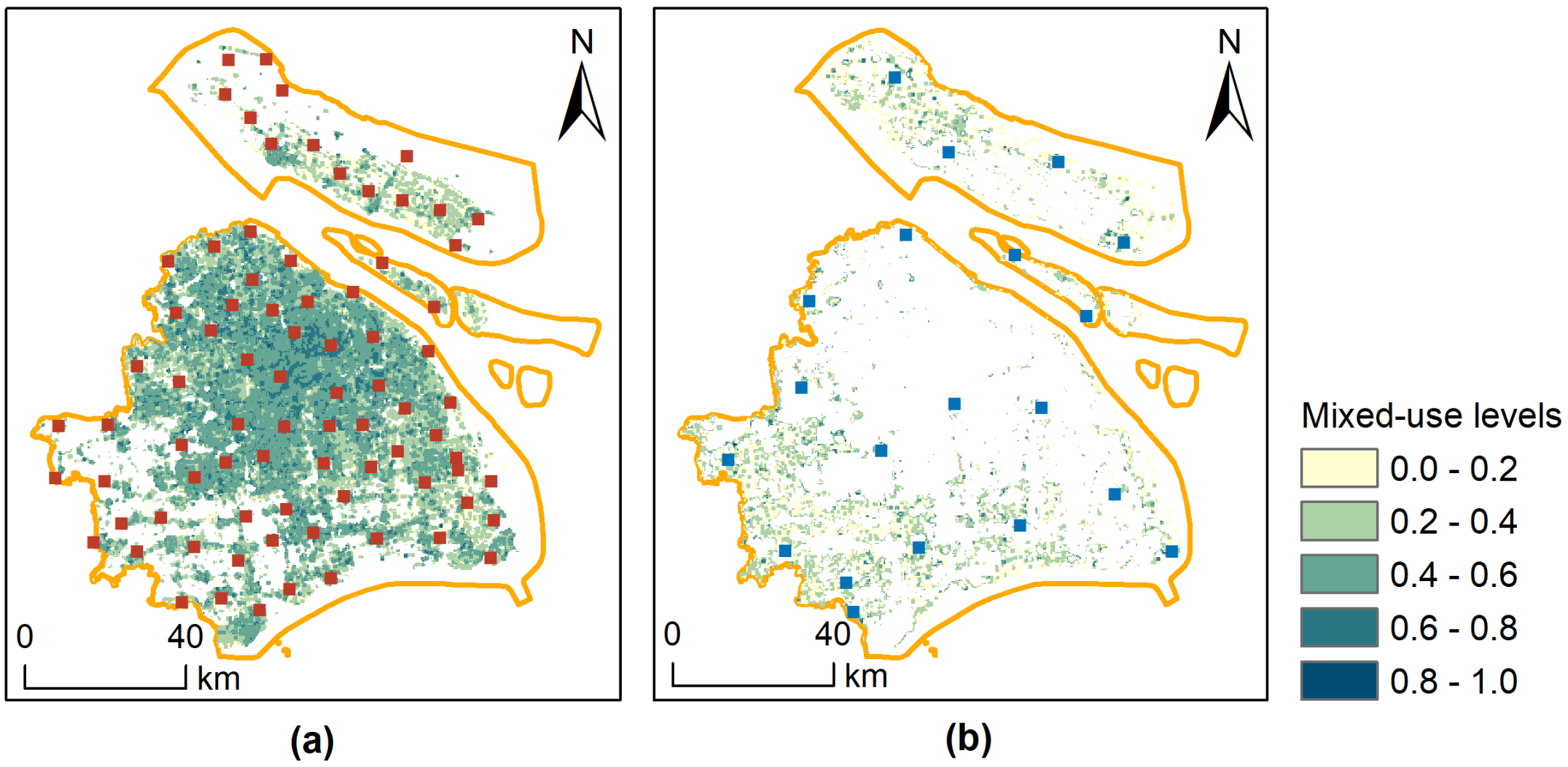

**Fig. 9.** Spatial distribution of optimal samples. (a) Optimal sample locations (marked in orange) within the building-dense stratum. (b) Optimal sample locations (marked in blue) within the building-sparse stratum.

To validate the efficacy of our proposed sampling method, we conduct a hands-on inspection of the chosen image samples. Some of these are presented in Fig. 10. A clear spatial distinction is evident between the building-dense and building-sparse strata. The former prominently showcases a higher concentration of buildings compared to the latter. The high diversity of building entities across these samples stems from the optimization of mixed-use levels in our sampling technique. Even within the building-sparse stratum, most samples indicate a sporadic presence of buildings, rather than a complete absence. Given the stochastic nature of the simulated annealing optimization process, a few samples might not align with the described characteristics. However, this deviation doesn't undermine the overall sample quality and representativeness. Such a diverse sampling outcome equips the MT-BR to grasp varied building characteristics, especially in geographically complex contexts.

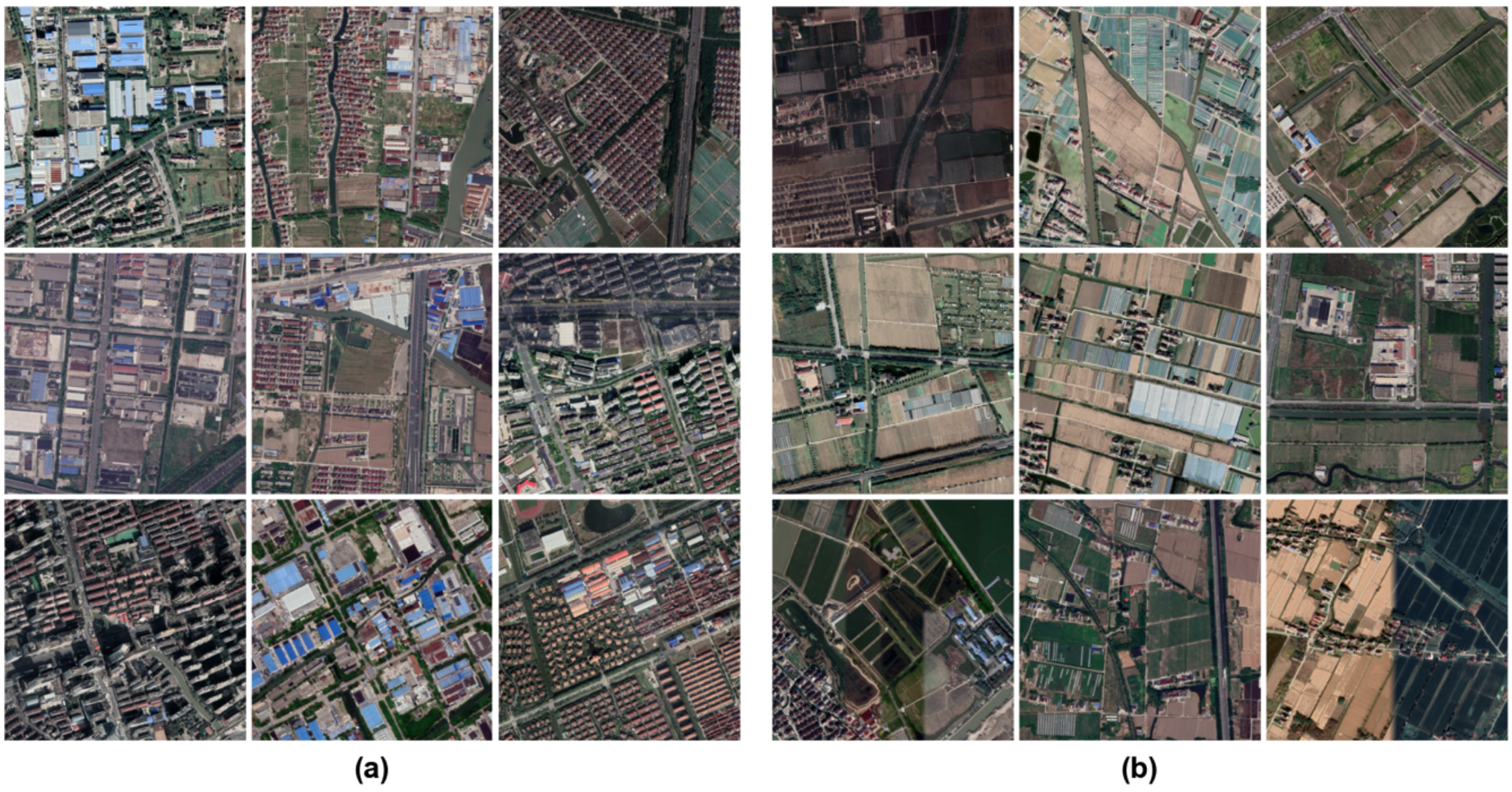

**Fig. 10.** A selection from the optimal image samples. (a) Image samples within the building-dense stratum. (b) Image samples within the building-sparse stratum.

In recent study, Sun et al. (2022) proposed a geospatial stratified and optimized sampling (GSOS) method, which emphasizes single-objective optimization centered on the spatial distribution of samples. The efficacy of GSOS was demonstrated to surpass traditional sampling methods, such as random and stratified sampling. Based on their work, this study introduces a dual-objective optimization-based sampling technique. To validate our method's effectiveness, we compare it against three other sampling strategies: random sampling, stratified sampling, and GSOS. We utilize four distinct sampling schemes to compile both training and testing datasets. Each training dataset comprises 500 sampling units, whereas each test dataset comprises 100 units. Moreover, to manage labeling costs effectively, this comparative research is primarily centered on rooftop extraction. Following Sun et al. (2022), we utilize the DeepLab V3+, training it on our assembled training datasets using a publicly available vectorized rooftop dataset for annotations (Z. Zhang et al., 2022). To ensure a balanced evaluation, networks trained with different sampling approaches are assessed against four distinct test datasets.

As illustrated in Fig. 11, networks trained on datasets derived from our sampling method exhibit a distinct superiority. When assessed against identical test datasets, these networks achieve superior F1 scores and IoU metrics in comparison to those trained with alternative datasets, presenting a margin of roughly 0.5% to 1%. This empirical assessment underscores the efficacy of our sampling technique in curating representative samples. Additionally, stratified sampling and methods that consider the optimization of spatial distribution appear to offer advantages over naive random sampling in the selection of representative samples. The outcomes also provide compelling evidence of the pivotal role such representative samples hold in effectively training deep neural networks. Consequently, these networks can achieve enhanced predictive accuracy without necessitating modifications to sophisticated network architectures.

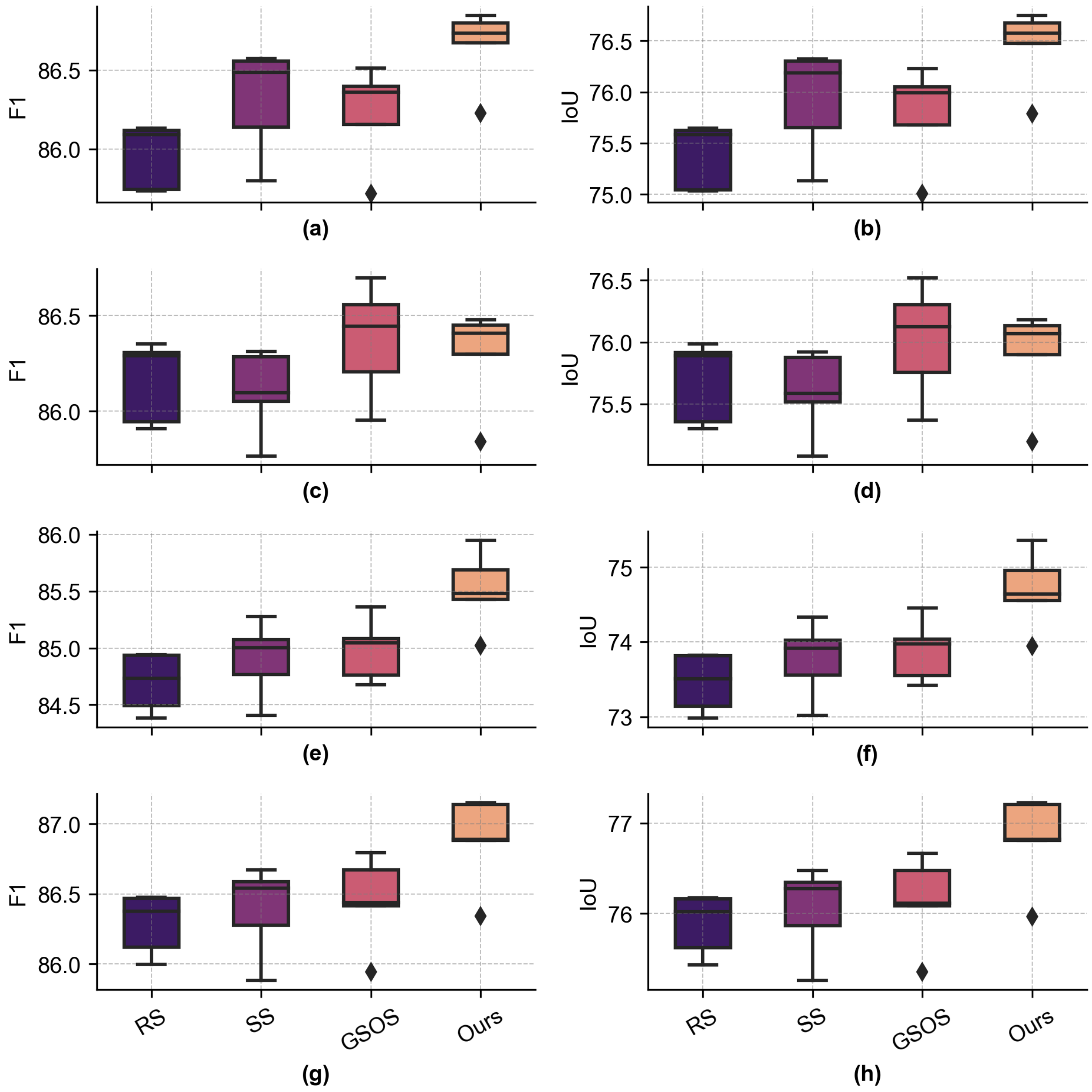

**Fig. 11.** Comparative performance of various sampling methods, represented by F1-score and IoU metrics across different test datasets. Each row highlights results from a distinct test dataset, and individual boxes denote performance statistics of DeepLab V3+ trained with different training datasets. (a-b) Results tested on datasets from random sampling (RS). (c-d) Results tested on datasets from stratified sampling (SS). (e-f) Results tested on datasets employing geospatial stratified and optimized sampling (GSOS). Notably, GSOS, as proposed by Sun et al. (2022), utilizes single-objective optimization focusing on the spatial distribution of samples. (g-h) Results tested on datasets derived from our sampling scheme.

### *5.1.3 Data pre-processing*

Upon curating representative samples, the subsequent phase involves labeling and annotation activities, which include image slicing and dataset segmentation. For this task, we employ ArcGIS Pro for manual annotations of rooftops within the image data and the identification of roof architectural types. Using the reclassified AOIs data, rooftops were matched to determine the urban functional type of each building. For rooftops without a direct overlay from the AOIs, the urban functional type was manually determined. Given our model's requirement for an input dimension of 512×512 pixels, both the primary images and their associated labels were processed through mask cropping, utilizing a consistent 512×512 pixel sliding window. This process generated 900 segmented image patches. Of these, 80% were designated for training, 10% for validation, and the remaining 10% for testing. Fig. 12 shows that the distribution of building details is consistent across the training, validation, and testing datasets.

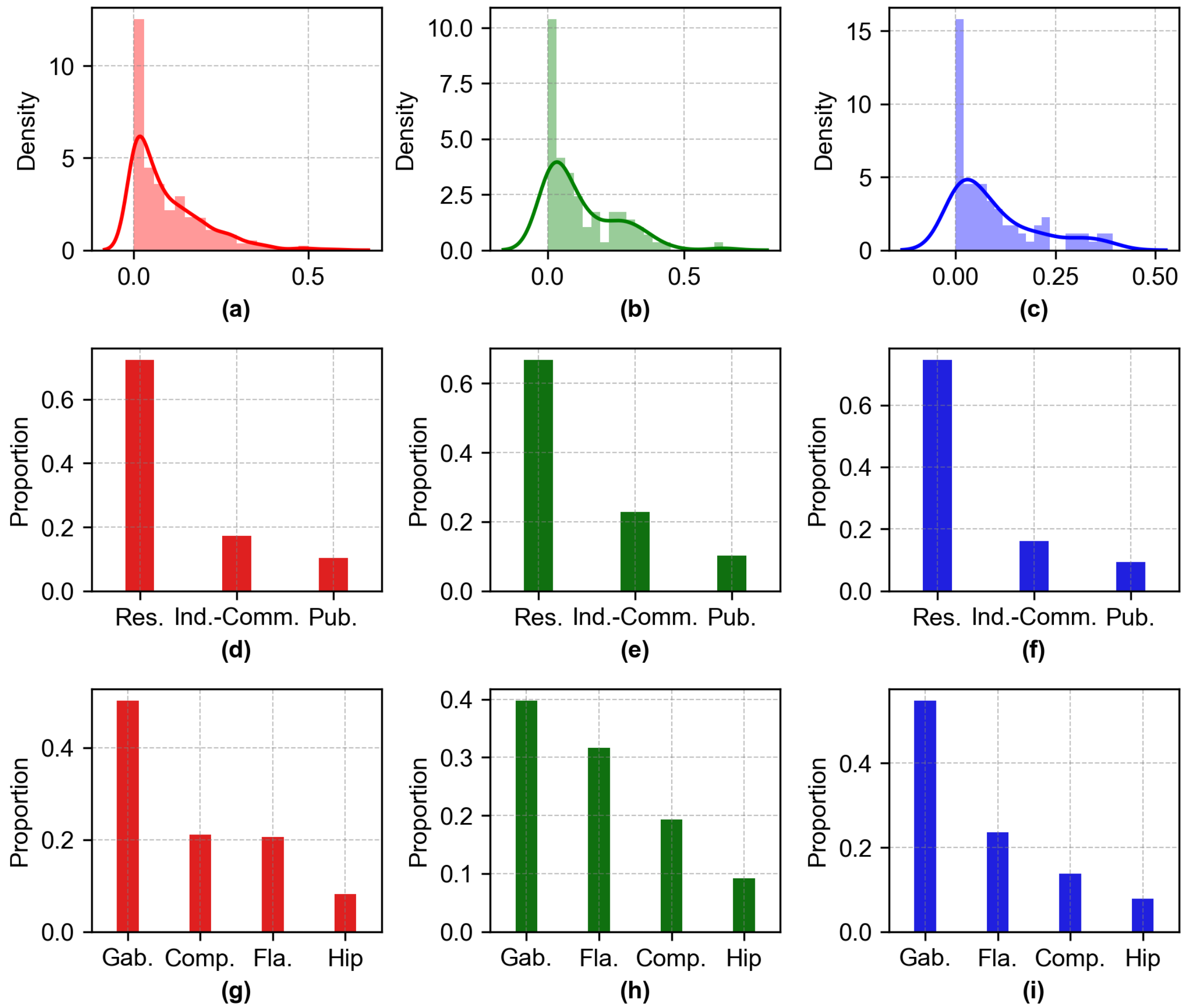

**Fig. 12**. Distribution of annotated building details across the datasets. The red graphs denote the training dataset, green graphs correspond to the validation dataset, and blue graphs represent the test dataset. (a-c) Rooftop density distribution across the respective datasets. (d-f) Urban functional type distribution across the respective datasets. (g-i) Roof architectural type distribution across the respective datasets.

### *5.2 Comparative and ablation experiments*

In order to evaluate the effectiveness of the proposed methods, we conduct a series of comparison and ablation experiments using our optimally selected dataset. This encompasses both qualitative and quantitative analyses, with an emphasis on the consistency of predicted outcomes, as well as the precision and efficiency of inferences. The methods involved in the comparative evaluation encompass five networks dedicated to semantic segmentation tasks: UNet (Ronneberger et al., 2015), DeepLab V3+ (Chen et al., 2018), UPerNet (Xiao et al., 2018), SegFormer (Xie et al., 2021), and SETR (Zheng et al., 2021), as well as five networks designed for instance segmentation tasks: Mask RCNN (He et al., 2017), MS R-CNN (Huang et al., 2019), HTC (Chen et al., 2019a), SOLO V2 (Wang et al., 2020), and Mask2Former (Cheng et al., 2022). This diverse set of networks encompasses a broad range of cutting-edge techniques in visual recognition, ranging from convolutional neural network-based to Transformer-based designs, and spanning both single and dual-stage methodologies. A concise overview of these comparative methods is presented below.

- UNet (Ronneberger et al., 2015) is a renowned architecture for image segmentation, and features a U-shaped network structure designed to effectively capture contextual information.
- DeepLab V3+ (Chen et al., 2018) is distinguished by its deep convolutional neural network equipped with atrous convolutions, allowing it to handle large receptive fields and intricate details.
- UPerNet (Xiao et al., 2018) focuses on multi-scale feature integration and refinement, enhancing the representation of objects in various sizes within an image.
- SegFormer (Xie et al., 2021) introduces Transformer architecture to the segmentation task and is recognized for its strong performance in capturing global contextual information. For our experiments, we employ the robust SegFormerB5 model.
- SETR (Zheng et al., 2021) captures long-range dependencies and global context in images by leveraging the self-attention mechanism, addressing intricate spatial correlations and fine-grained details effectively.
- Mask RCNN (He et al., 2017) is a two-stage instance segmentation method that combines object detection with pixel-wise mask prediction, making it suitable for precise object localization.

- MS R-CNN (Huang et al., 2019) extends the Mask RCNN by addressing multi-scale challenges, improving the handling of objects at different sizes and resolutions.
- HTC (Chen et al., 2019a), or Hybrid Task Cascade, enhances instance segmentation by sequentially refining object masks, resulting in better accuracy.
- SOLO V2 (Wang et al., 2020) is a single-stage instance segmentation model, eliminating the need for multi-stage processing while maintaining high accuracy and speed.
- Mask2Former (Cheng et al., 2022) is a Transformer-based model for instance segmentation, directly predicting object masks and classes, efficiently handling complex scenes without region proposals.

Furthermore, a straightforward approach to enhance the capability of single-task networks to handle multiple learning tasks is by merging tasks to simultaneously manage the extraction of various attributional details. For instance, identifying urban functional types can be structured as a tri-class classification task, while recognizing roof architectural types can be approached as a quad-class classification task. Merging these tasks culminates in a composite task featuring a twelve-class classification. Subsequently, the predictions can be deconstructed to yield results for discrete tasks. We incorporate this mixed-class strategy within the instance segmentation methods, thereby establishing pertinent experimental benchmarks.

### *5.2.1 Analysis of result consistency*

Initially, we examine the results produced by different approaches to understand their unique characteristics, using Mask R-CNN, DeepLab V3+, Mask R-CNN with the mixed-class strategy, and MT-BR as representative examples. When employing Mask R-CNN and DeepLab V3+ to extract diverse building details, separate sequential runs are required, resulting in inconsistencies in their respective prediction outcomes. These discrepancies are evident in the red boxes of Fig. 13, showcasing predictions from Mask R-CNN and DeepLab V3+. In contrast, the mixed-class methodology and our proposed MT-BR adeptly sidestep this inconsistency issue. Furthermore, rooftops derived using the semantic segmentation method, such as DeepLab V3+, tend to merge into a single entity. This contrasts with the outputs of the instance segmentation method, Mask R-CNN, which delineates distinct and countable instances. This unique capability of instance segmentation method influenced our choice to adopt it as a baseline method.

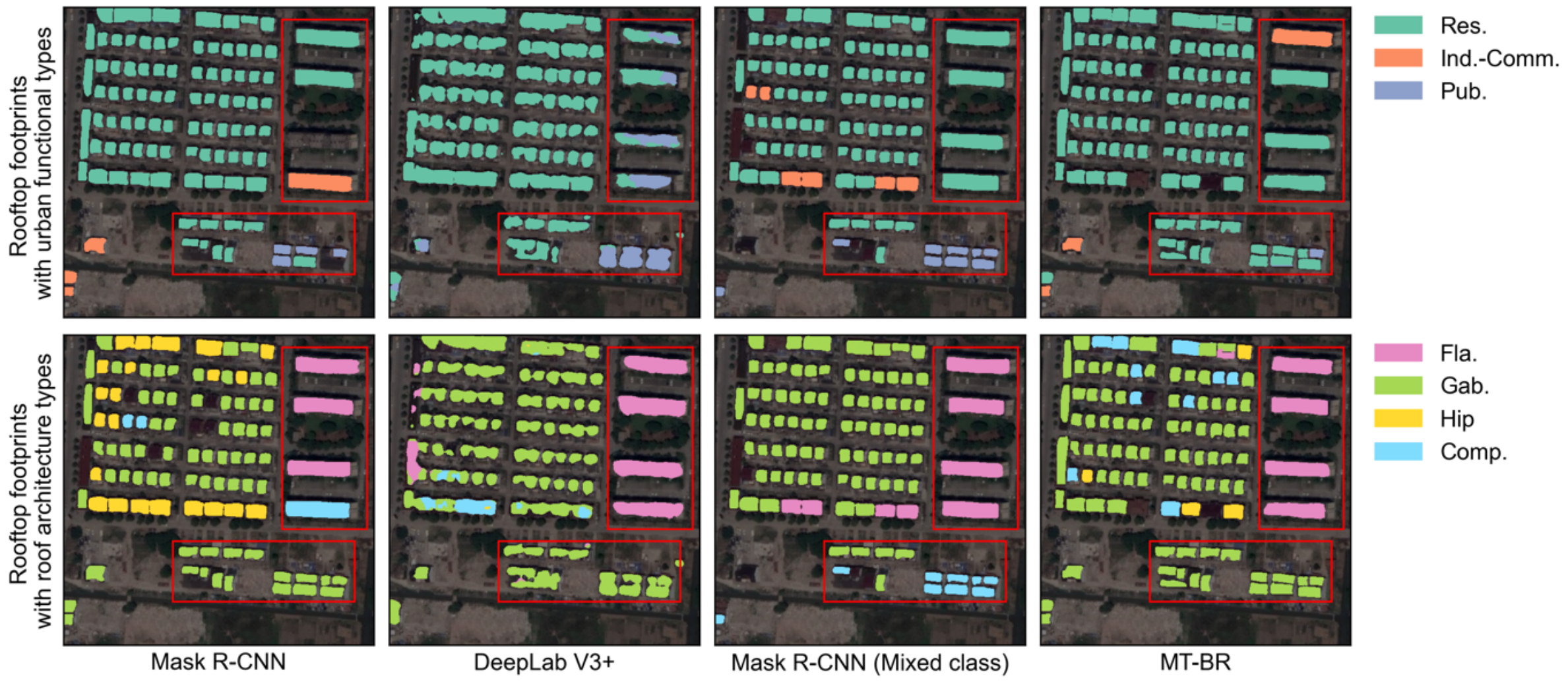

**Fig. 13.** Depiction of prediction result consistency.

#### *5.2.2 Assessment of prediction accuracy*

We conduct a series of ablation experiments to validate the architectural design and augmentation strategies of MT-BR, with the quantitative findings presented in Tables 2 and 3. The outcomes underscore the efficacy of adopting both fully-connected and convolutional approaches in multi-branches, as well as the utilization of deformable convolution in bolstering prediction accuracy for both spatial and attributional information extraction. Additionally, the tailored augmentation strategies further enhance MT-BR's predictive accuracy. When compared with multi-model decision fusion, test-time augmentation exhibits a decline in prediction accuracy. Such results shed light on the potential significance of image scales and positioning in the extraction of various building details from high-resolution satellite imagery. Post-processing not only offers improvements in the accuracy of building attributional information extraction, but its implementation is also crucial to ensuring consistent pixel types within individual building instances. Consequently, subsequent experiments and practical applications in Shanghai employ the proposed augmentation strategies, excluding test-time augmentation.

**Table 2** Ablation analysis for prediction accuracy of MT-BR's architecture. The baseline embodies the extended Mask R-CNN with multi-branches. FC-Conv denotes the incorporation of fully-connected and convolutional layers in multi-branches. DCN stands for deformable convolutional network. A higher score (↑) signifies better performance.

| Baseline | FC-Conv branches | DCN | Rooftop | | Urban functional type | Roof architectural type |
|---|---|---|---|---|---|---|
| | | | F1 (↑) | IoU (↑) | Kappa (↑) | Kappa (↑) |
| ✓ | | | 78.89 (± 1.01) | 65.14 (± 1.38) | 69.60 (± 0.70) | 61.90 (± 0.84) |
| ✓ | ✓ | | 79.07 (± 0.56) | 65.39 (± 0.76) | 69.56 (± 0.76) | 62.33 (± 0.14) |
| ✓ | | ✓ | 79.67 (± 0.54) | 66.22 (± 0.74) | 70.20 (± 0.72) | 63.08 (± 0.83) |
| ✓ | ✓ | ✓ | **80.18 (± 0.48)** | **66.91 (± 0.66)** | **70.72 (± 0.71)** | **64.47 (± 0.82)** |

**Table 3** Ablation analysis for prediction accuracy of augmentation strategies. TTA represents test time augmentation, while MMDF indicates multi-model decision fusion. A higher score (↑) signifies better performance.

| MT-BR | HRNet | Ensemble Strategies | | Post-processing | Rooftop footprint | | Urban functional type | Roof architecture type |
|---|---|---|---|---|---|---|---|---|
| | | TTA | MMDF | | F1 (↑) | IoU (↑) | Kappa (↑) | Kappa (↑) |
| ✓ | | | | | 80.18 (± 0.48) | 66.91 (± 0.66) | 70.72 (± 0.71) | 64.47 (± 0.82) |
| ✓ | ✓ | | | | 82.03 (± 0.49) | 69.54 (± 0.71) | 71.87 (± 0.62) | 67.20 (± 0.46) |
| ✓ | ✓ | ✓ | | | 81.50 (± 0.70) | 68.78 (± 0.99) | 71.76 (± 0.36) | 65.12 (± 1.04) |
| ✓ | ✓ | | ✓ | | 84.30 (± 0.14) | 72.86 (± 0.21) | 74.57 (± 0.38) | 69.81 (± 0.32) |
| ✓ | ✓ | ✓ | ✓ | | 83.95 (± 0.13) | 72.34 (± 0.19) | 74.21 (± 0.09) | 68.71 (± 0.39) |
| ✓ | ✓ | | ✓ | ✓ | **84.30 (± 0.14)** | **72.86 (± 0.21)** | **74.67 (± 0.59)** | **70.04 (± 0.37)** |

We subsequently compare our proposed methods with selected state-of-the-art techniques, with the results delineated in Table 4. Given the inconsistencies in rooftops extracted by single-task methods, we average the evaluation metrics for their separate predictions. The results reveal that MT-BR, especially when enhanced with augmentation strategies, showcases marked superiority over other methods in both spatial and attributional information extraction. While the mixed-class approach can handle multiple tasks simultaneously, its accuracy in attributional information extraction is notably lower. This underperformance might be attributed to a pronounced class imbalance. Specifically, upon examining the attributes of labeled building instances, we observe an imbalanced distribution, as illustrated in Fig. 14. The distributions of urban functional types and roof architectural types are skewed, and their combined representation further amplifies this imbalance, posing challenges for deep neural networks. Moreover, the comparison underscores that methods designed for semantic segmentation tasks underperform in spatial information extraction compared to their counterparts designed for instance segmentation. This insight offers a valuable reference for real-world applications. It's also noteworthy that sophisticated methods based on Transformer, such as SegFormer, SETR, and Mask2Former, have not achieved as remarkable results as they might in other general datasets. This could be attributed to the fact that training these intricate models typically demands a large number of datasets, which are limited in our study. Additionally, when deploying these methods for multi-task operations, tailored designs and adjustments could be essential to harness their full potential.

**Table 4** Assessment of prediction accuracy. A higher score (↑) signifies better performance.

| Methods | Rooftop | | Urban functional type | Roof architectural type |
|---|---|---|---|---|
| | F1 (↑) | IoU (↑) | Kappa (↑) | Kappa (↑) |
| UNet | 71.11 (± 1.11) | 55.22 (± 1.34) | 63.47 (± 0.60) | 51.49 (± 0.82) |
| DeepLab V3+ | 77.11 (± 0.58) | 62.76 (± 0.77) | 68.83 (± 0.65) | 61.10 (± 1.42) |
| UPerNet | 77.55 (± 0.76) | 63.35 (± 1.01) | 69.65 (± 0.90) | 62.88 (± 0.70) |
| SegFormer | 72.77 (± 0.80) | 57.23 (± 0.98) | 66.03 (± 0.99) | 54.73 (± 1.67) |
| SETR | 63.10 (± 0.82) | 46.13 (± 0.86) | 56.45 (± 0.72) | 45.09 (± 1.13) |
| Mask R-CNN | 78.85 (± 0.56) | 65.11 (± 0.76) | 69.25 (± 1.08) | 62.59 (± 1.29) |
| MS R-CNN | 78.56 (± 0.57) | 64.70 (± 0.76) | 68.96 (± 1.04) | 62.29 (± 0.54) |
| HTC | 75.96 (± 0.58) | 61.24 (± 0.75) | 65.00 (± 0.90) | 57.82 (± 0.81) |
| SOLOv2 | 74.20 (± 0.65) | 59.01 (± 0.81) | 65.32 (± 1.36) | 56.96 (± 1.56) |
| Mask2Former | 77.16 (± 0.85) | 62.83 (± 1.13) | 69.38 (± 1.39) | 61.98 (± 1.27) |
| Mask R-CNN (Mixed class) | 77.43 (± 0.93) | 63.18 (± 1.23) | 60.88 (± 0.73) | 48.47 (± 0.86) |
| MS R-CNN (Mixed class) | 77.45 (± 1.23) | 63.21 (± 1.65) | 60.19 (± 1.60) | 47.06 (± 1.36) |
| HTC (Mixed class) | 69.42 (± 3.65) | 53.26 (± 4.29) | 53.38 (± 4.01) | 41.32 (± 2.72) |
| SOLO V2 (Mixed class) | 69.10 (± 0.97) | 52.79 (± 1.13) | 54.01 (± 1.20) | 43.73 (± 1.02) |
| Mask2Former (Mixed class) | 72.27 (± 1.49) | 56.60 (± 1.82) | 56.94 (± 1.56) | 45.10 (± 1.14) |
| MT-BR | **80.18 (± 0.48)** | **66.91 (± 0.66)** | **70.72 (± 0.71)** | **64.47 (± 0.82)** |
| MT-BR (Augmentation) | **84.30 (± 0.14)** | **72.86 (± 0.21)** | **74.67 (± 0.59)** | **70.04 (± 0.37)** |

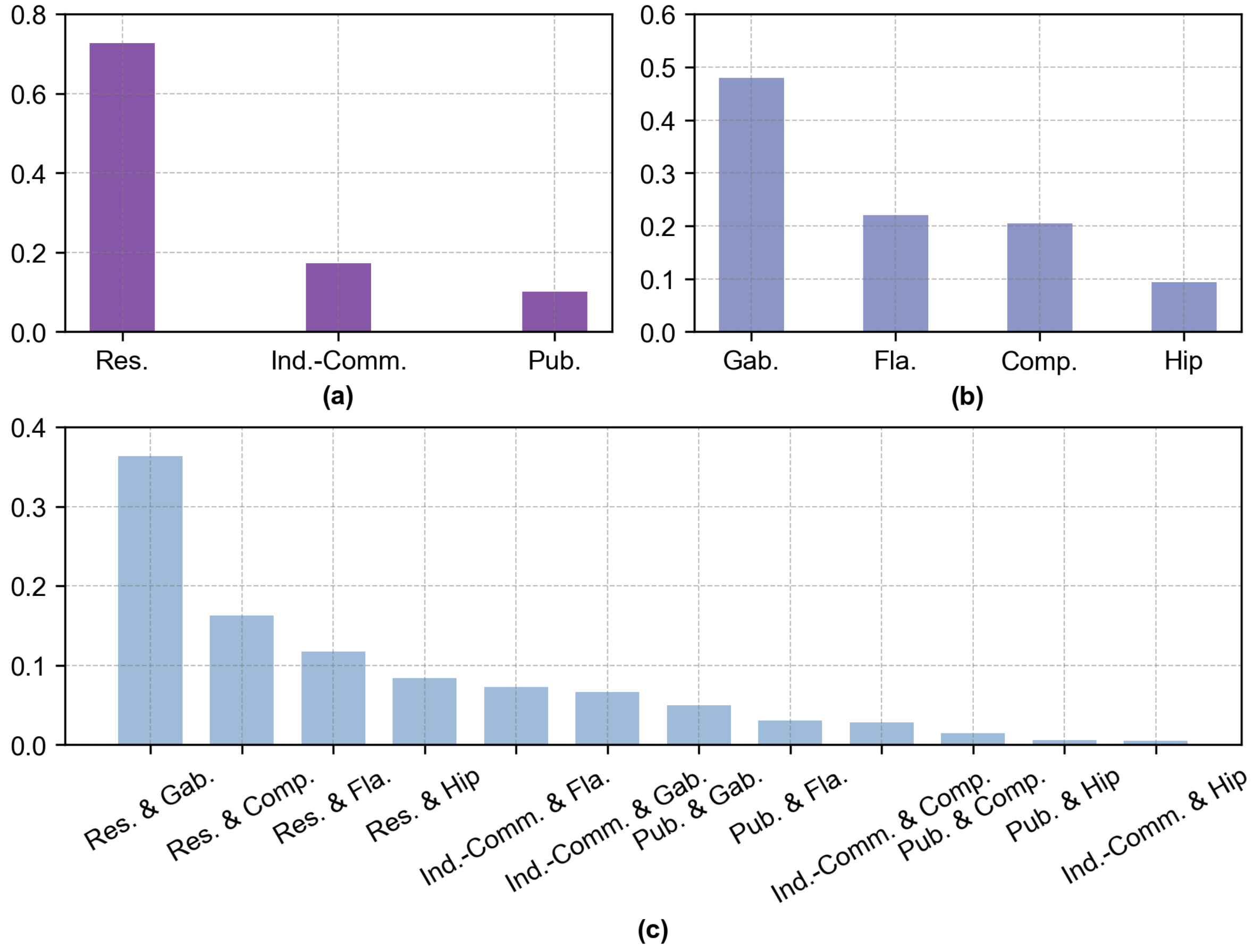

**Fig. 14.** Distribution of building instance attributes. (a) Distribution of urban functional types. (b) Distribution of roof architectural types. (c) Combined distribution of both types.

Subsequently, we undertake a qualitative assessment of the outputs from various methods, as depicted in Figs. 15-17. For this analysis, we chose a mix of urban and rural images, showcasing varying building densities, ranging from highly dense areas to sparsely populated ones. The visualized predictions are consistent with our earlier quantitative observations. Our proposed method demonstrates commendable precision and recall in identifying building rooftops. It successfully identifies buildings often missed by other methods, all the while upholding a high accuracy. While the proposed approach occasionally misclassifies certain attributes, its outputs are largely congruent with the ground truth, underscoring its robust feature representation and inference capabilities. However, the semantic segmentation methods encounter difficulties in accurately defining specific buildings, often resulting in an amalgamation of their outcomes, as evident in Fig. 15. In contrast, instance segmentation-based methods excel in demarcating individual structures and correlating them with the appropriate attributes, as illustrated in Fig.16. Similar to the quantitative evaluation, instance segmentation methods employing the mixed-class strategy reveal certain limitations, such as missing building entities and misclassifying attributes, especially in densely populated images, as seen in Fig.17. Through empirical evaluations across diverse geographical settings, the MT-BR, enhanced with augmentation techniques, consistently showcases superior prediction capabilities and robust generalization.

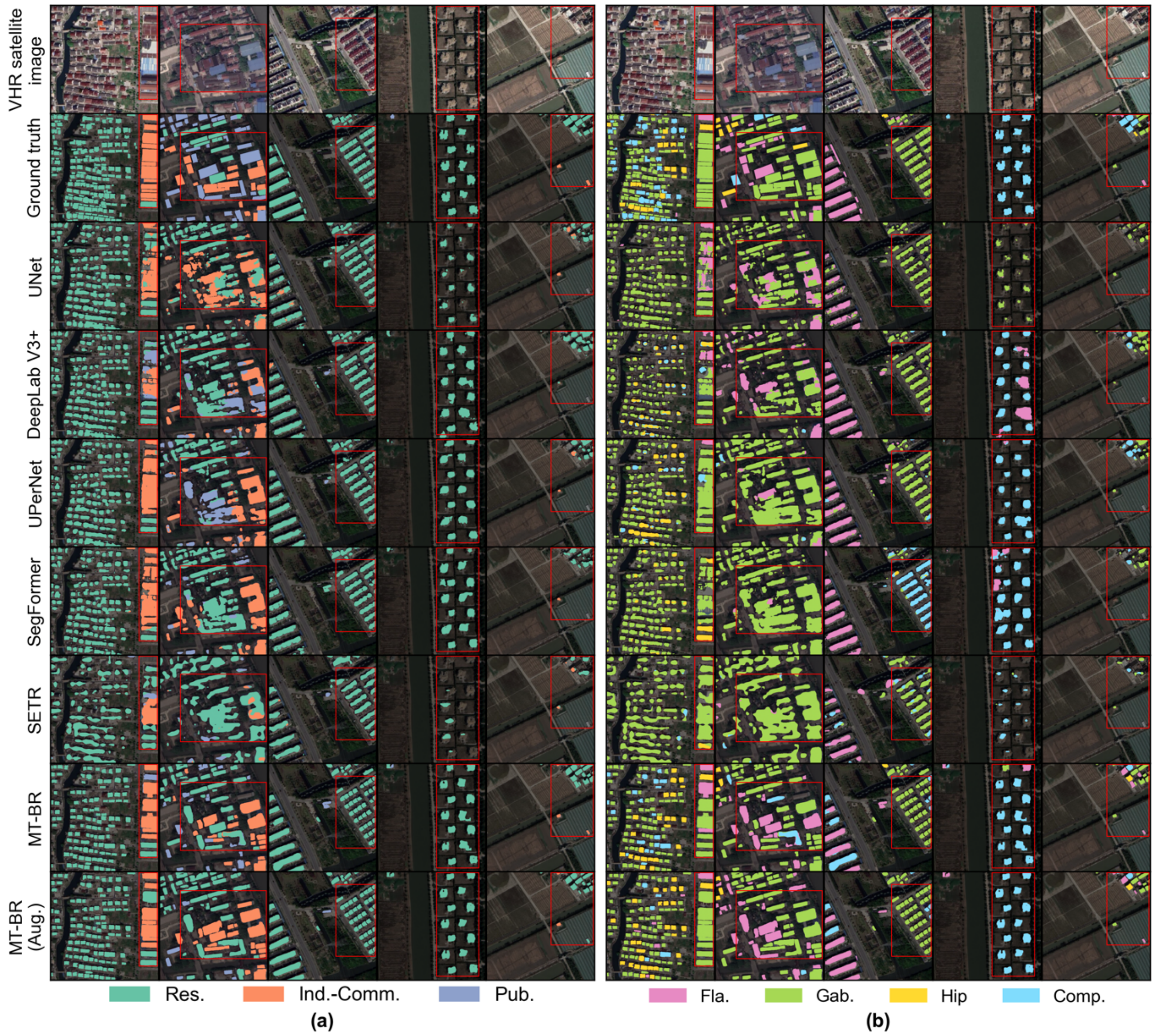

**Fig. 15.** Predictive visualization from semantic segmentation methods. (a) Urban functional types associated with rooftops. (b) Roof architectural types associated with rooftops. “Aug.” is the prediction augmentation.

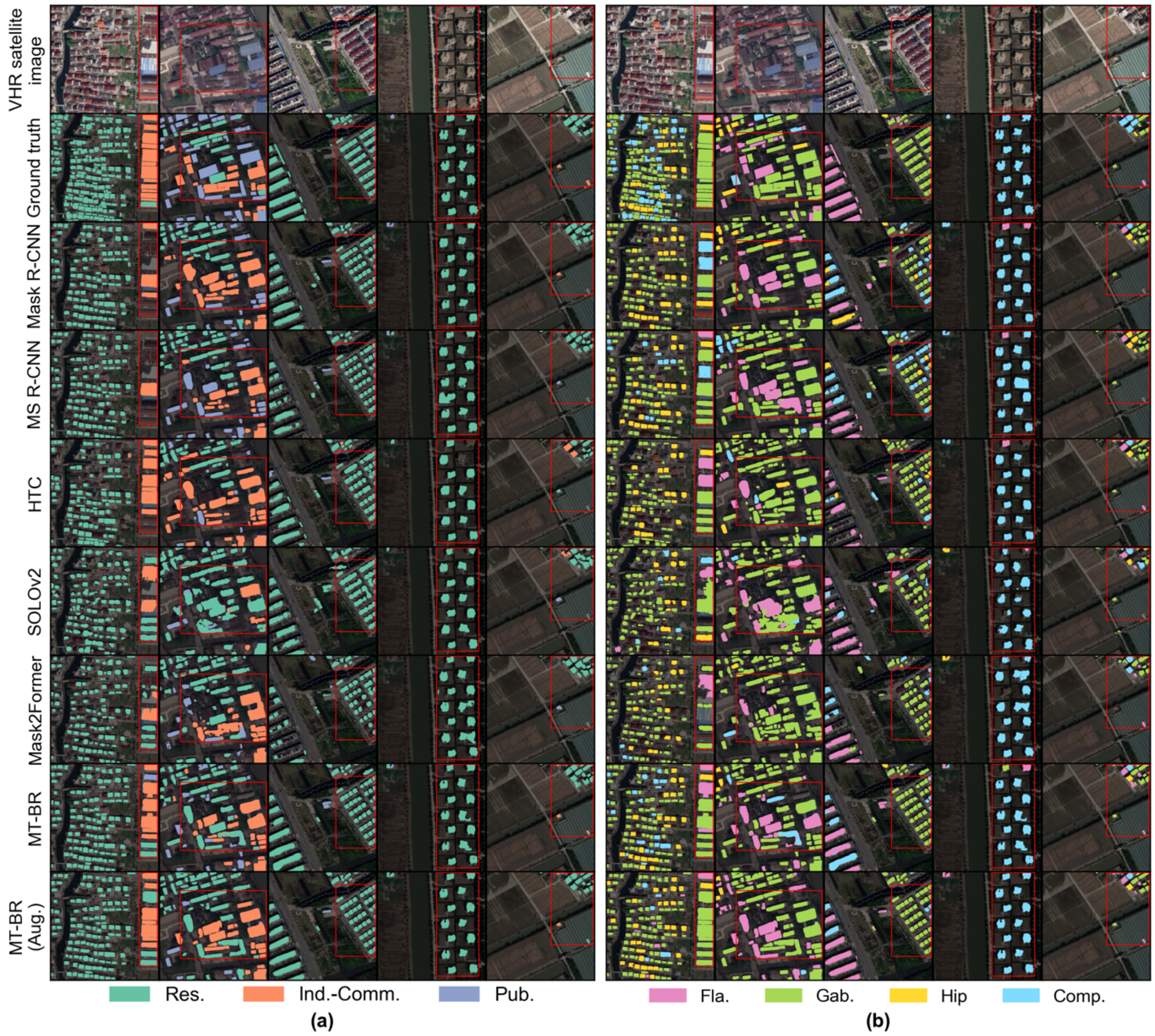

**Fig. 16.** Predictive visualization from instance segmentation methods. (a) Urban functional types associated with rooftops. (b) Roof architectural types associated with rooftops. “Aug.” is the prediction augmentation.

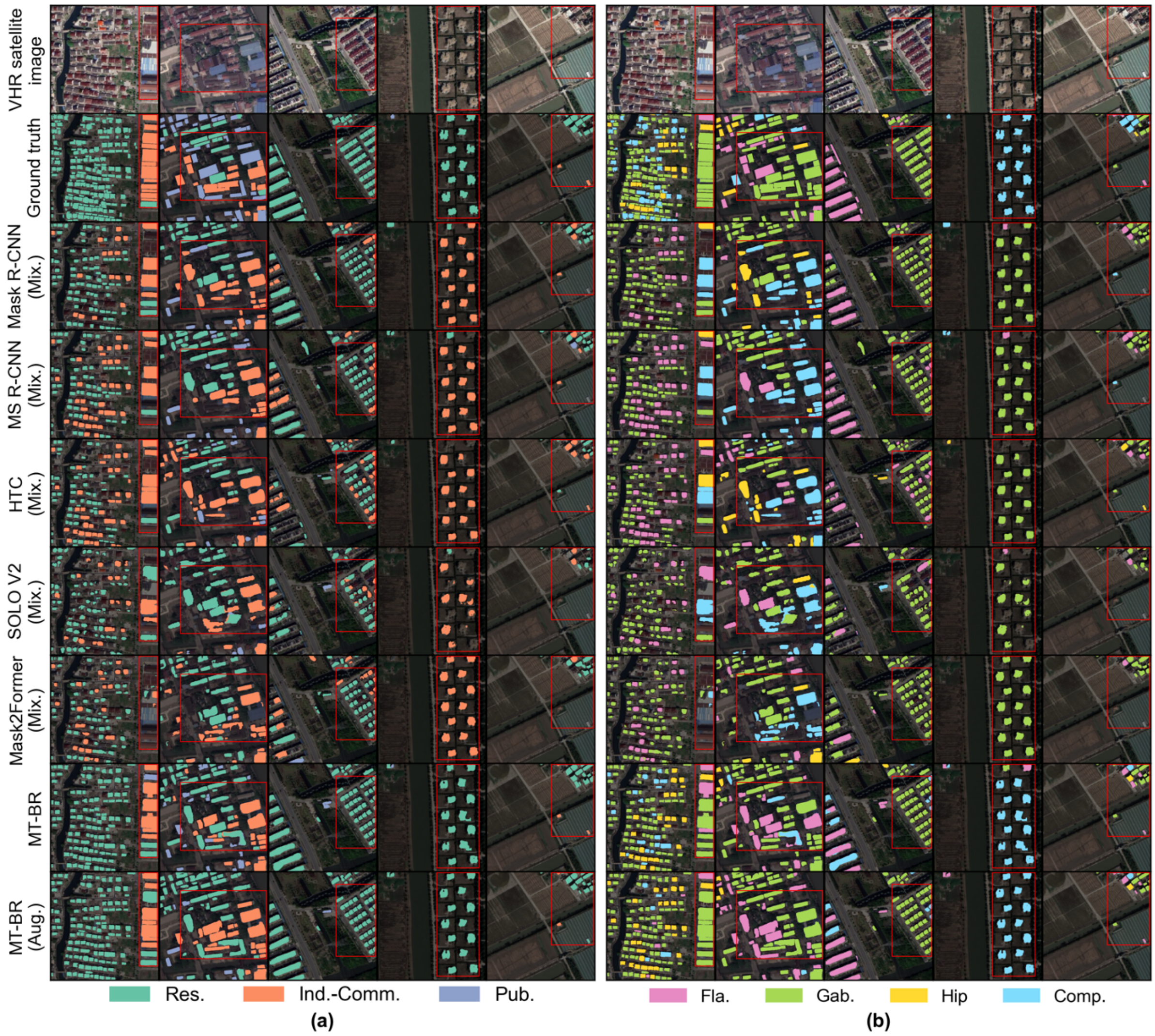

**Fig. 17.** Predictive visualization from instance segmentation methods utilizing the mixed-class strategy. (a) Urban functional types associated with rooftops. (b) Roof architectural types associated with rooftops. "Mix." is the mixed-class strategy, and "Aug." is the prediction augmentation.

### *5.2.3 Assessment of inference efficiency*

The extraction of various building information requires the single-task methods to run through multiple workflows, rather than in an end-to-end manner. Consequently, we assess the inference efficiency of the single-task methods, ideally calculating time costs by summing the durations needed for different tasks. The efficiency comparisons, tabulated in Table 5, highlight that employing a mixed-class strategy significantly bolsters multi-tasking efficiency, resulting in the fastest inference times. It's worth noting that methods based on semantic segmentation prove to be more time-efficient than their instance segmentation counterparts. This discrepancy arises from the inherently intricate structures of instance segmentation methods, designed to tackle both detection and segmentation tasks concurrently.

While MT-BR boasts an end-to-end design and inference capability, its speed does not markedly surpass that of mixed-class strategy methods. To delve deeper, we conducted an ablation study on MT-BR's inference efficiency, with findings summarized in Table 6. Several factors can impede the achievement of optimal inference speeds, the most significant one of which is the simultaneous utilization of fully-connected and convolutional layers across multiple branches. Furthermore, the incorporation of deformable convolution can slightly reduce the inference duration. By avoiding from utilizing the methodology that combines fully-connected and convolutional operations in multi-branches, the inference speed of MT-BR outperforms the majority of the chosen comparative methods, while yet maintaining a commendable level of prediction accuracy (see Table 2 and Table 4). These experimental findings provide prospective users with valuable insights, enabling them to make informed decisions on deploying MT-BR based on their specific requirements—whether they prioritize swift computation or heightened prediction accuracy.

**Table 5** Assessment of inference efficiency. Lower inference times (↓) and higher FPS scores (↑) signifies faster processing speed.

| Methods | Inference times per image / ms (↓) | FPS (↑) |
|---|---|---|
| UNet | 53.10 (± 0.40) | 18.83 (± 0.14) |
| DeepLab V3+ | 69.78 (± 0.26) | 14.33 (± 0.05) |
| UPerNet | 76.44 (± 0.66) | 13.08 (± 0.11) |
| SegFormer | 117.99 (± 0.61) | 8.48 (± 0.04) |
| SETR | 126.82 (± 0.76) | 7.89 (± 0.05) |
| Mask R-CNN | 80.99 (± 0.20) | 12.35 (± 0.03) |
| MS R-CNN | 84.58 (± 0.38) | 11.82 (± 0.05) |
| HTC | 146.02 (± 1.91) | 6.85 (± 0.09) |
| SOLOv2 | 82.91 (± 0.26) | 12.06 (± 0.04) |
| Mask2Former | 221.34 (± 0.81) | 4.52 (± 0.02) |
| Mask R-CNN (Mixed class) | **40.75 (± 0.26)** | **24.54 (± 0.16)** |
| MS R-CNN (Mixed class) | 43.11 (± 0.30) | 23.20 (± 0.16) |
| HTC (Mixed class) | 84.74 (± 1.21) | 11.80 (± 0.17) |
| SOLO V2 (Mixed class) | 41.71 (± 0.21) | 23.98 (± 0.12) |
| Mask2Former (Mixed class) | 108.96 (± 0.47) | 9.18 (± 0.04) |
| MT-BR | 119.04 (± 1.56) | 8.40 (± 0.11) |

**Table 6** Ablation analysis for inference efficiency of MT-BR. Lower inference times (↓) and higher FPS scores (↑) signifies faster processing speed.

| Baseline | FC-Conv branches | DCN | Inference times per image / ms (↓) | FPS (↑) |
|---|---|---|---|---|
| ✓ | | | **57.61 (± 0.23)** | **17.36 (± 0.07)** |
| ✓ | ✓ | | 111.61 (± 0.60) | 8.96 (± 0.05) |
| ✓ | | ✓ | 67.39 (± 0.21) | 14.84 (± 0.05) |
| ✓ | ✓ | ✓ | 119.04 (± 1.56) | 8.40 (± 0.11) |

### *5.3 Generalization assessment and large-scale application*

In the final phase, we employ the MT-BR equipped with augmentation strategies to extract comprehensive building details across the entirety of Shanghai. We implement the expansion prediction technique to address the challenges posed by uneven transitions during the stitching of image patches. The originally rasterized dataset is converted to the ESRI Shapefile format and streamlined using the Douglas-Peucker algorithm for the sake of data management and transmission. Fig. 18 showcases the extracted building details across Shanghai.

The generated dataset reveals approximately 1.77 million building entities, closely aligning with the number of buildings detailed in the 2020 publicly available dataset for Shanghai, which registers around 1.72 million buildings (Z. Zhang et al., 2022). The delineated building rooftops mirror the administrative boundary of Shanghai, with the city's urban core displaying prominent building clusters, indicative of Shanghai's robust economic development. When compared to the background data provided by the high-resolution satellite imagery, the extracted building entities demonstrate a notable level of generality and precision. The success of this work can be attributed to our sampling method that ensured the inclusion of representative data samples as input, and the development of MT-BR with augmentation strategies, which effectively managed multiple recognition tasks. The approach finely distinguishes between individual buildings and intricate geographical contexts, ensuring the accurate delineation of rooftops across both urban and rural settings.

Moreover, each building is discernibly mapped with roof architectural types, as seen in Fig. 18 (a), and urban functional types as in Fig. 18 (b). This mapping aligns coherently with Shanghai's current development and planning. A significant portion of the buildings, 48.40%, have gable roofs, while flat roofs constitute 32.31%.

Urban central areas predominantly showcase flat-roofed buildings, whereas rural or suburban regions lean towards gable roofs. Functionally, residential buildings dominate the landscape, making up 76.78%, while industrial-commercial and public facility buildings account for 18.93% and 4.29%, respectively. Among residential building, gable roofs are prevalent, accounting for 50.83%, with flat and complex roof types each covering around 20%. On the other hand, commercial, industrial, and public facility buildings primarily exhibit flat and gable types. It is worth noting that the dataset generated in this work is openly available to the public and holds significant promise for advancing the understanding of urban dynamics and fostering sustainable development.

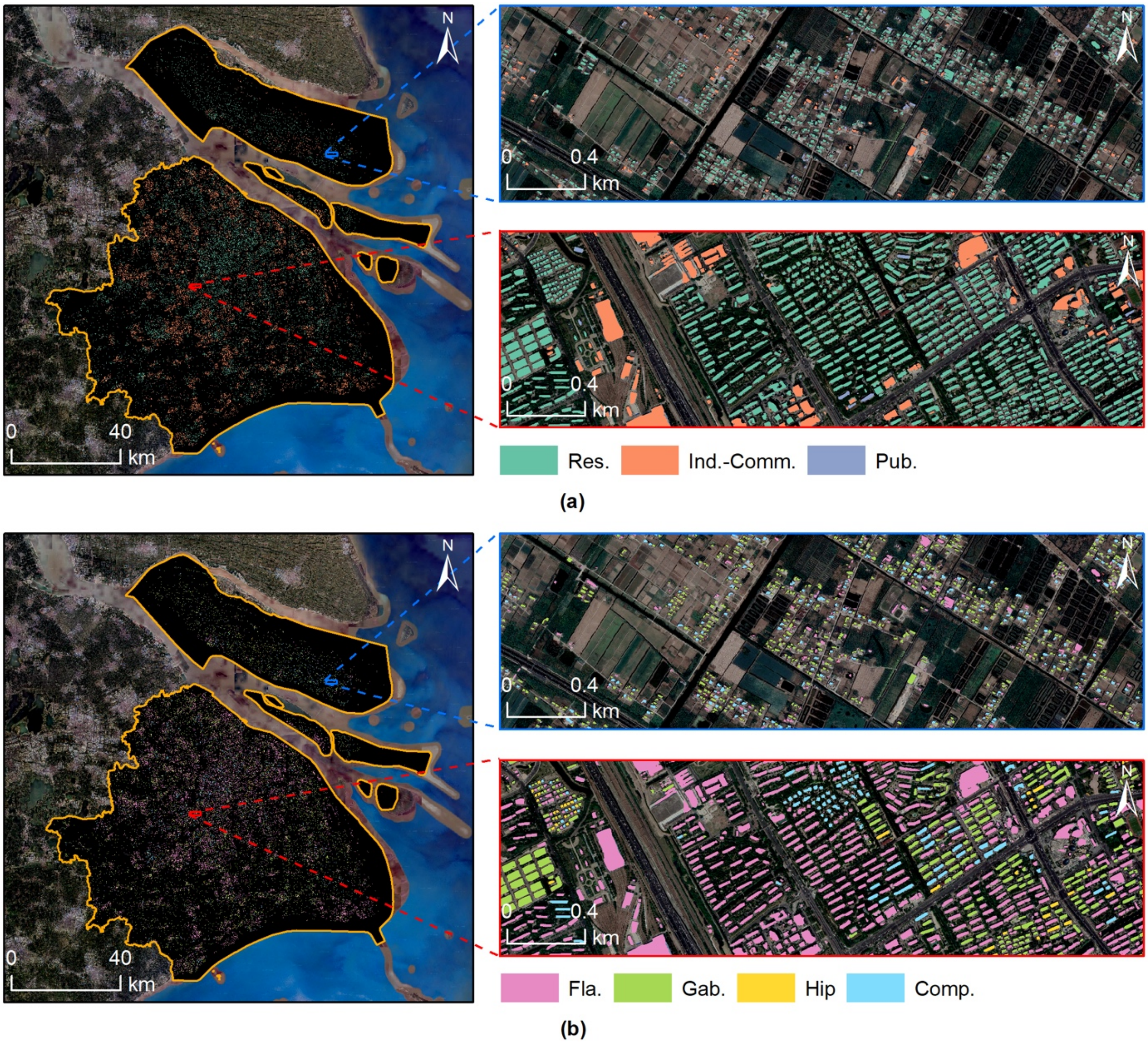

**Fig. 18.** Building details delineated in Shanghai. (a) Urban functional types associated with rooftops. (b) Roof architectural types associated with rooftops.

## 6. Conclusion

This study presented the MT-BR, an approach designed to address the limitations of single-task methodologies. Focusing on building details like rooftops, urban functional types, and architectural roof types, we validated the utility of MT-BR. Its flexible design allows for potential extensions to capture diverse building information from high-resolution satellite imagery. For large-scale applicability, we developed a dual-objective optimized spatial sampling scheme. This scheme prioritizes both the spatial distribution of samples and the urban characteristics they represent, aiming to select limited but representative training samples and enhance MT-BR's predictive performance. The effectiveness of this approach is confirmed by an empirical experiment. Subsequently, we introduce augmentation strategies to refine MT-BR's capabilities. In comparative tests, MT-BR shows consistent improvements over other deep learning methods, with performance gains ranging from 2% to 20%, further amplified by 5% with our augmentation techniques. Our ablation studies also provide guidance for users, allowing them to customize MT-BR's architecture based on their specific requirements. At last, the application of our methodologies in Shanghai demonstrates their practical utility and the quality of the generated building-related datasets.

Despite the advancements presented in this study, there remain avenues for further research. While high-resolution satellite imagery offers detailed physical insights, it sometimes falls short in providing a comprehensive view of building information, especially when discerning subtle attribute types. A more holistic understanding of urban environments might be achieved by incorporating diverse data sources, such as social media feeds or multi-sensor satellite imagery. With the emerging prominence of Transformer-based methods tailored for multi-modal data, future efforts could explore these methodologies. However, they necessitate vast datasets and judicious architectural design. Furthermore, the estimation of building height, an essential parameter, poses its own set of challenges and will be a central focus in our forthcoming research endeavors.

## Author contribution

**Zhen Qian:** Conceptualization, Methodology, Software, Writing - Original Draft, Writing - Review & Editing. **Min Chen:** Conceptualization, Methodology, Supervision, Writing - Review & Editing, Funding acquisition. **Zhuo Sun:** Methodology, Writing – review & editing. **Fan Zhang:** Methodology, Writing – review & editing. **Qingsong Xu:** Methodology, Formal analysis, Writing – review & editing. **Jinzhao Guo:** Investigation. **Zhiwei Xie:** Writing – review & editing. **Zhixin Zhang:** Formal analysis, Validation, Writing – review & editing.

## Declaration of competing interest

The authors declare that they have no known competing financial interests or personal relationships that could have appeared to influence the work reported in this paper.

## Data availability

The primary code and dataset obtained from this work, containing building details from Shanghai, are available for public access at the following repository: https://github.com/ChanceQZ/BuildingDetails-Multitask.

## Acknowledgments

This work was supported in part by the National Natural Science Foundation of China (No. 42325107), in part by the National Natural Science Foundation of China (No. 42101353), and in part by the Humanities and Social Sciences Foundation of the Ministry of Education of China (General Program) (No. 21YJC790129).